\documentclass[runningheads]{llncs}

\usepackage{eccv}

\usepackage{eccvabbrv}

\usepackage{graphicx}
\usepackage{booktabs}
\usepackage{amssymb}
\usepackage{pifont}
\usepackage[accsupp]{axessibility}  

\usepackage{orcidlink}

\newcommand{\newpara}[1]{\vspace{3pt}\noindent\textbf{#1}}

\usepackage[normalem]{ulem}
\usepackage{wrapfig}
\usepackage{multirow}
\usepackage{color, colortbl}
\usepackage{tikz}
\usetikzlibrary{tikzmark, arrows.meta}
\usepackage{verbatim}
\usepackage{tcolorbox}

\definecolor{Gray}{gray}{0.93}
\newcolumntype{g}{>{\columncolor{Gray}}c}

\definecolor{Blue}{rgb}{0.9, 0.95, 1}
\newcolumntype{b}{>{\columncolor{Blue}}c}
\newcommand{\cmark}{\ding{51}}
\newcommand{\xmark}{\ding{55}}

\begin{document}

\title{Recognizing Co-Speech Gestures in-the-Wild}

\titlerunning{GRW}

\author{Sindhu B Hegde \and
K R Prajwal \and
Andrew Zisserman}

\authorrunning{S Hegde et al.}

\institute{Visual Geometry Group, Dept. of Engineering Science, University of  Oxford\\
\email{\{sindhu,prajwal,az\}@robots.ox.ac.uk}\\ 
\url{https://www.robots.ox.ac.uk/~vgg/research/grw}}

\maketitle

\begin{abstract}
While humans naturally gesture during speech, only a sparse subset of these co-speech gestures are visually depictive and semantically linked to specific spoken words.  In this paper, we introduce a large-scale dataset -- {\em Gesture Recognition in the Wild} (GRW), comprising co-speech gestures corresponding to a diverse vocabulary of 155 words. GRW contains 140k manually annotated video clips where the word is spoken, with 17k instances of semantic co-speech gestures including their frame-level temporal boundaries. The video clips are collected `in the wild' from public-facing discourse, including lectures, talk shows, and interviews, covering a diverse range of speakers and visual conditions.

We also introduce video models to: (a) classify gestures as semantic or not; (b) recognize the word corresponding to a co-speech gesture; and (c) temporally localize the gesture. These models are trained and evaluated on the GRW dataset and compared against a range of strong baselines, establishing benchmark results for all three tasks. The dataset, annotations, and trained models are publicly available on the project website.

\end{abstract}

\section{Introduction}
\label{sec:intro}

\begin{figure}
    \centering
    \vspace{-30pt}
    \includegraphics[width=\linewidth]{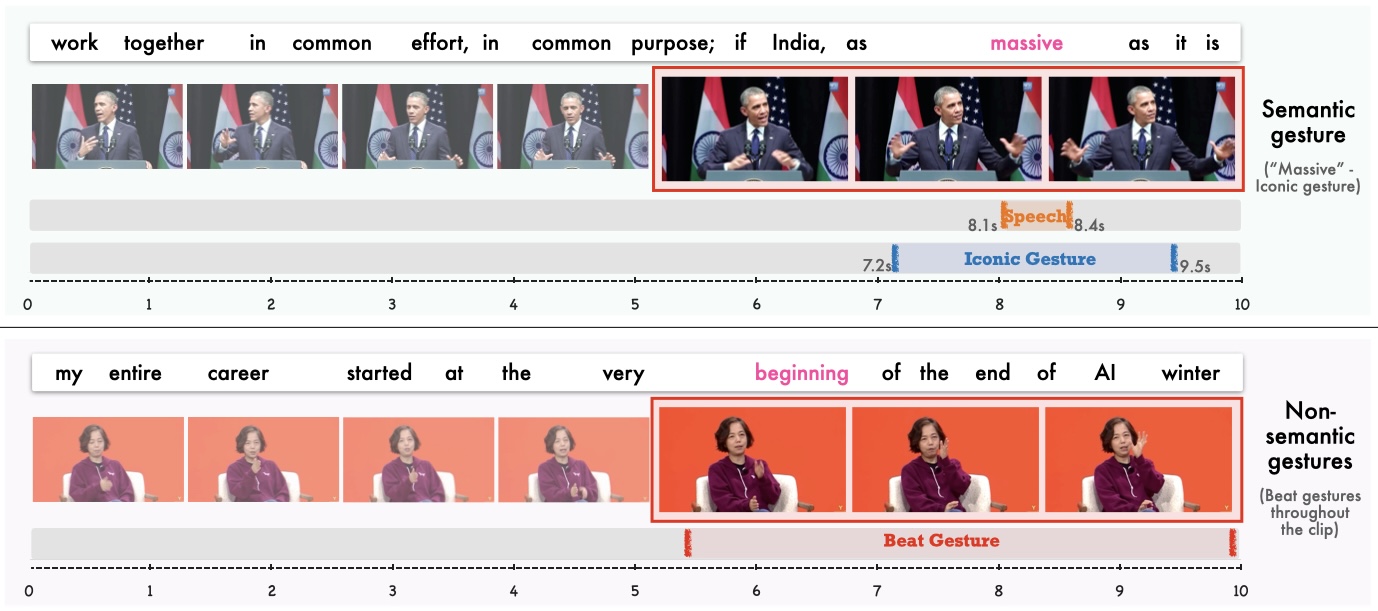}
    \vspace{-15pt}
    \caption{Semantic vs. non-semantic co-speech gestures. \textbf{(Top)} Obama performs an iconic semantic gesture for the word ``massive''. Notice the significant temporal offset between the physical gesture and the spoken word. \textbf{(Bottom)} There are no semantic gestures around the word ``beginning''. This paper focuses on building dataset and models that can automatically recognize and localize semantic gestures in real-world clips.}
    \label{fig:teaser}
    \vspace{-30pt}
\end{figure}

Humans naturally move their hands and upper body while speaking, yet only a sparse subset of these movements are \emph{meaning-bearing}. In this work, we focus on \textbf{semantic co-speech gestures}, i.e.\ instances where a speaker produces a clear, visually depictive (iconic, deictic, or metaphoric) movement directly attributable to a specific spoken word (e.g., tracing a \emph{circle} while saying `circle', or moving the hands \emph{apart} while saying `massive' in Fig~\ref{fig:teaser}). Humans also make other gestures while speaking, such as \emph{beat} gestures (rhythmic emphasis) and incidental motions (fidgeting, resting hands), but these are considered \emph{non-semantic}.

Unfortunately, no large-scale dataset currently exists that has annotations for these semantic co-speech gestures in unconstrained situations. Distinguishing these semantic gestures in-the-wild is challenging: they are relatively sparse in time, highly variable in form, and often require integrating both linguistic and visual context.
Consequently, while recent multimodal foundation models for video understanding~\cite{zhu2024languagebind,lin2024videollava,wang2024internvideo2} excel at generic recognition and retrieval, they remain remarkably brittle when tasked with recognizing specific, fine-grained gestures. To address this critical gap, we make the following contributions:

\noindent \textbf{Gesture Recognition in the Wild (GRW) Dataset}: We introduce the first large-scale, unconstrained benchmark for semantic co-speech gestures. GRW features manually curated annotations for $\approx$ 140k video clips across 37k unique identities, indicating whether or not a semantic gesture is present. For $\approx$17k semantic instances, we provide word labels and frame-accurate temporal boundaries for both the speech and the corresponding gesture, explicitly accounting for the natural temporal misalignment between the two. The word labels are distributed across a vocabulary of 155 conceptual words.

\noindent \textbf{Insights into dynamics of gesture and spoken words}: Leveraging the diversity of our data, we provide a comprehensive analysis of how humans gesture in the wild. We reveal fundamental insights into the temporal envelope of gestures (is the word gestured before it is spoken or after?) and the relative likelihood of words to elicit a semantic gesture (which words are most often gestured?)
    
\noindent \textbf{New models for three core gesture tasks}: We formalize three core gesture understanding tasks: (i) semantic gesture classification -- is the gesture semantic or not? (ii) word-level gesture recognition, and (iii) temporal gesture localization. We then develop and train models for these tasks using the GRW dataset. Our results demonstrate the importance of modeling extended temporal motion context for semantic gesture classification. Finally, we show that our models outperform existing state-of-the-art vision-language models, establishing a strong benchmark and foundation for future multimodal gesture research.

\vspace{-10pt}
\section{Related Work}
\label{sec:related_work}

\vspace{-5pt}
Human gestures are studied across multimodal machine learning, HCI, graphics, and linguistics. Existing resources vary significantly in gesture type (isolated commands vs.\ co-speech vs.\ sign language), sensing modality (RGB/RGB-D, 2D/3D pose), and supervision granularity. 

\begin{table}[t]
\centering
\caption{Comparison of gesture datasets. The Gesture Recognition in the Wild (GRW) dataset uniquely provides verified semantic word labels and precise temporal gesture boundaries for in-the-wild co-speech gestures while supporting multimodal inputs.}
\vspace{-10pt}
\label{tab:dataset_comparison}
\scriptsize 
\setlength{\tabcolsep}{1pt}
\begin{tabular}{l | l | c | c | c | c | c}
\toprule
\textbf{Dataset} & \textbf{Modality} & \begin{tabular}{@{}c@{}}\textbf{\#}\\\textbf{Identities}\end{tabular} & \textbf{Hours} & \begin{tabular}{@{}c@{}}\textbf{In-the-}\\\textbf{wild?}\end{tabular} & \begin{tabular}{@{}c@{}}\textbf{Gestured}\\\textbf{word labels}\end{tabular} & \begin{tabular}{@{}c@{}}\textbf{Temporal}\\\textbf{Bounds}\end{tabular} \\
\midrule

\rowcolor{Gray}
\multicolumn{7}{l}{\textbf{Isolated / Command-Style}} \\
Jester~\cite{materzynska2019jester} & RGB & 1,376 & $\sim$123 & \xmark & \xmark & \cmark \\
EgoGesture~\cite{zhang2018egogesture} & RGB-D & 50 & 24 & \xmark & \xmark & \cmark \\
NVGesture~\cite{molchanov2016online} & RGB-D+IR & 20 & $\sim$10 & \xmark & \xmark & \cmark \\
ChaLearn LAP~\cite{wan2016chalearn} & RGB-D & 21 & $\sim$14 & \xmark & \xmark & \cmark \\

\midrule
\rowcolor{Gray}
\multicolumn{7}{l}{\textbf{Co-Speech Gestures}} \\
Trinity~\cite{ferstl2018investigating} & Mocap+Audio & 1 & 4 & \xmark & \xmark & \xmark \\
TalkingWithHands~\cite{lee2019talking}& Mocap+Audio & 50 & 150 & \xmark & \xmark & \xmark \\
BEAT~\cite{liu2022beat} & Mocap+Audio & 30 & 76 & \xmark & \cmark & \xmark \\
Speech2Gesture~\cite{ginosar2019learning} & 2D+Audio & 10 & 144 & \cmark & \xmark & \xmark \\
PATS~\cite{ahuja2020no} & 2D+Audio+Text & 25 & 251 & \cmark & \xmark & \xmark \\
AVS-Spot~\cite{hegde2025understanding} & RGB+Audio+Text & 384 & 0.4 & \cmark & \cmark & \xmark \\
SeamlessInteraction~\cite{agrawal2025seamless} & RGB+Audio+Text & 4,000 & 4,065 & \xmark & \xmark & \xmark \\
\midrule
\rowcolor{Blue}
\textbf{Ours (GRW)} & \textbf{RGB+Audio+Text} & 37,409 & 173 & \textbf{\cmark} & \textbf{\cmark} & \textbf{\cmark} \\
\bottomrule
\end{tabular}%
\vspace{-15pt}
\end{table}

\newpara{Gesture Datasets.}
\label{subsec:rw_datasets}
Many datasets focus on predefined, isolated command gestures (e.g., swipes, thumbs-up) for HCI applications. Examples include Jester (27 classes)~\cite{materzynska2019jester}, EgoGesture (83 classes)~\cite{zhang2018egogesture}, NVGesture~\cite{molchanov2016online}, SocialGesture~\cite{cao2025socialgesture} (4 deitic gestures in multi-person interaction settings), and the ChaLearn LAP challenges (IsoGD/ConGD) scaling to hundreds of categories~\cite{wan2016chalearn}. While valuable, these datasets model gestures-as-commands and lack the spoken-word grounding and linguistic alignment necessary for co-speech semantic modeling.

\vspace{-10pt}
\paragraph{Sign language datasets:} Corpora like BOBSL~\cite{albanie2021bobsl}, How2Sign~\cite{duarte2021how2sign}, WLASL~\cite{li2020word}, and YouTube-ASL~\cite{uthus2023youtube} benchmark sign language recognition and generation. Here, hand shape and motion are the primary communication channel rather than an auxiliary speech accompaniment. Although annotated at the word level, their discrete linguistic structure differs fundamentally from co-speech gestures, which are optional, highly variable, and only partially aligned with lexical items, preventing the direct transfer of methods.  

\vspace{-10pt}
\paragraph{Co-speech gesture datasets:} High-fidelity 3D motion capture datasets, such as Trinity Speech-Gesture~\cite{ferstl2018investigating} and Talking With Hands~\cite{lee2019talking}, provide clean trajectories, but lack diverse speakers and contexts. The BEAT dataset~\cite{liu2022beat} adds emotion and semantic-relevance annotations, categorizing beats, iconic, deictic, and metaphoric gestures. But, these datasets are limited by scale and diversity. A recent large-scale dataset, Seamless Interaction~\cite{agrawal2025seamless} is massive in scale ($4000+$ speakers, $4000+$ hours) but it is recorded in laboratory settings where dialogues are enacted with instructions to the actors, and do not contain gestured word labels either. There are also ``in-the-wild'' datasets used in gesture synthesis works, that source videos from TED talks and TV shows. For example, Speech2Gesture~\cite{ginosar2019learning} ($144$ hours, $10$ speakers) and PATS~\cite{ahuja2020no} ($251$ hours, $25$ speakers) aggregate auto-detected poses, audio, and transcripts. While critical for scaling generation models, these datasets lead to models that do not generate semantic gestures well, due to lack of specific annotations for the same~\cite{nyatsanga2023comprehensive}.

\newpara{Gesture Recognition and Understanding.}
Historically treated as an action recognition sub-problem, early gesture recognition employs 2D/3D CNNs and sequence models evaluated on isolated datasets~\cite{kopuklu2019real,crowdgestures}. Analyzing co-speech gestures, however, requires detecting their presence and segmenting temporal boundaries. A CRF-based sequence-labeling approach was proposed~\cite{ghaleb2024multiphase} to segment object-depicting gestures, and~\cite{ghaleb2024learning, ghaleb2025see} focused on learning gesture representations for face-to-face dialogues. However, these approaches were designed for constrained settings, rather than in-the-wild co-speech gestures. Only recently, JEGAL~\cite{hegde2025understanding} formalized tasks such as gesture-based retrieval and word spotting using tri-modal (video, speech, text) representations. While aligned with modeling semantics, these benchmarks focus on representation learning and retrieval rather than explicitly recognizing and segmenting semantic gestures with precise, word-level boundaries.

\newpara{Gesture Synthesis.}
Gesture generation has historically included rule-based and procedural systems inspired by linguistics. In recent years, data-driven models dominate, enabled by large gesture data and progress in sequence modeling. Speech2Gesture~\cite{ginosar2019learning} was a seminal work that learned personalized patterns from in-the-wild monologues using auto-extracted poses. Subsequent modern generative models like StyleGestures~\cite{alexanderson2020stylegestures} and ExpressGesture~\cite{ferstl2021expressgesture} improved realism, control, and coverage. Audio-driven models capture prosody well but struggle with semantic gestures. Incorporating transcripts or text embeddings has been a step towards addressing this. Rhythmic Gesticulator~\cite{ao2022rhythmic} structures explicit rhythm and semantics, while ConvoFusion~\cite{mughal2024convofusion} uses diffusion to emphasize specific words. Recent works like~\cite{liu2025semges} learns motion priors using VQ-VAEs for gesture synthesis, and~\cite{liu2026semconflow} generates gestures using a contrastive flow-matching framework. However, these models still predominantly rely on speech stress and prosody, noting that sparse semantic gestures remain difficult to capture due to a lack of non-beat annotations in training data.

\vspace{-10pt}
\section{The Gesture Recognition in the Wild (GRW) Dataset}
\label{sec:dataset}

\vspace{-30pt}
\begin{figure}
    \centering
    \includegraphics[width=\linewidth]{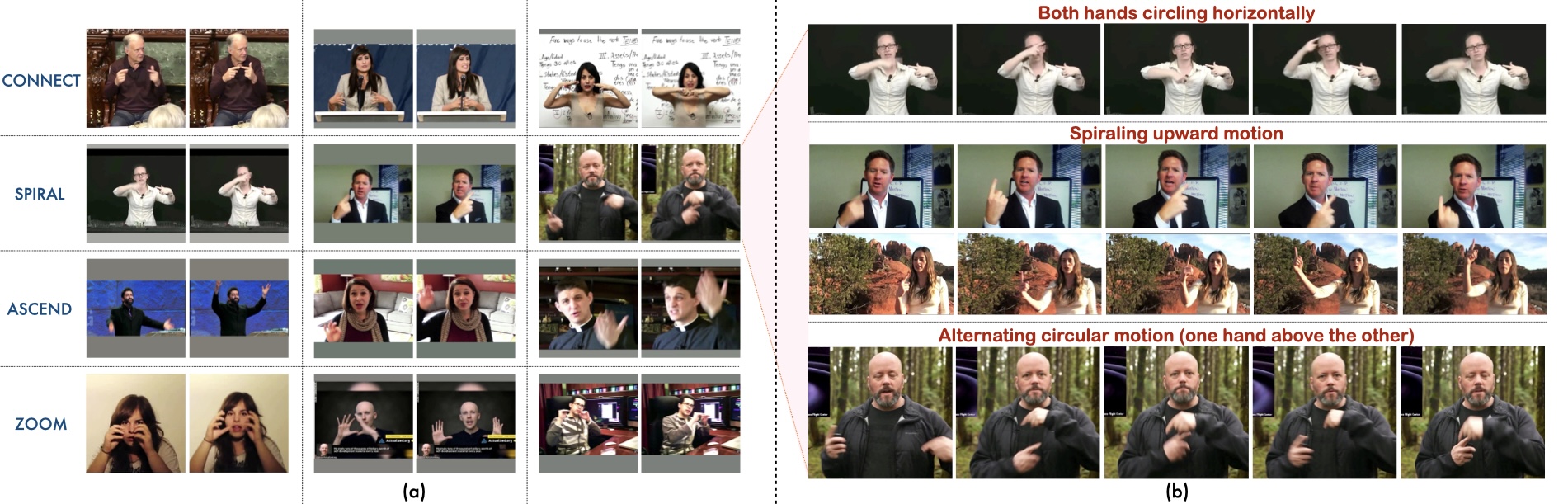}
    \vspace{-20pt}
    \caption{Samples from the \textbf{GRW} dataset. (a)  Lexical and kinematic diversity of semantic gestures. (b)  Even when gesturing the same conceptual word (``SPIRAL''), speakers employ different physical motions. The dataset captures these gestures across a wide variety of speakers, poses, and camera angles.}
    \label{fig:dataset}
    \vspace{-20pt}
\end{figure}

The GRW dataset comprises 139,503 manually annotated video clips curated from diverse, in-the-wild speaker environments. The core dataset features 17,340 precisely localized positive instances of semantic gestures, distributed across a broad vocabulary of 155 conceptual words. The dataset also comprises 122,163 verified negative samples where a target vocabulary word is spoken, but no corresponding semantic gesture is physically performed. Finally, we also include an automatically labeled set of 67,214 clips (refer to Sec~\ref{subsec:pretrain}) with word labels and approximate gesture boundaries that can be used for large-scale pre-training.
\begin{wrapfigure}{r}{0.45\textwidth}
    \vspace{-20pt}
    \centering
    \includegraphics[width=0.45\textwidth]{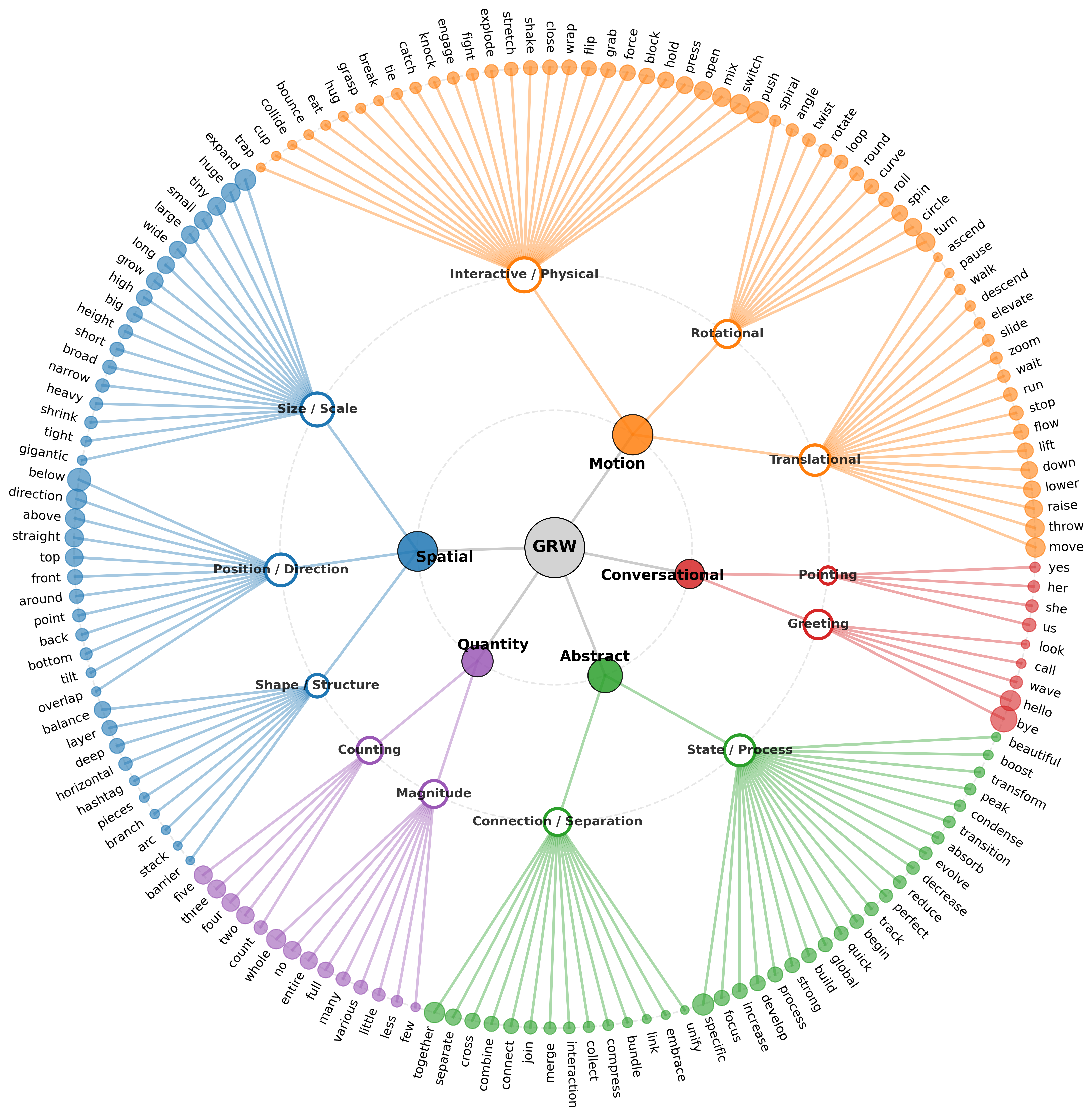}
    \caption{\textbf{Semantic Taxonomy of the GRW Dataset.} We organize our 155-word vocabulary into a three-tiered hierarchy, radiating from five high-level semantic domains to specific target words (leaf nodes). Node size is strictly proportional to the frequency of annotated clips per word.}
    \label{fig:gesture_taxonomy}
\end{wrapfigure}

\vspace{-10pt}
In Fig~\ref{fig:dataset}, we show a few representative samples from the dataset. Unlike laboratory-recorded isolated gesture datasets, GRW captures gestures in in-the-wild conditions public-facing discourse with natural intra- and inter-class variations. Fig.~\ref{fig:gesture_taxonomy} details the 155-word vocabulary across semantic domains ranging from literal actions (`Translational', Pointing') to complex concepts (`State/process', `Physical'). Node sizes (reflecting frequency) confirm a robust distribution across common anchors (e.g., `bye', `together') and nuanced descriptors (e.g., `spiral', expand'), establishing GRW as a comprehensive benchmark for real-world gesture understanding.
Each sample provides video, audio, word-level transcripts, and a binary semantic gesture label; positive samples additionally include the lemmatized target word $w$ and its precise temporal gesture boundary $\mathcal{B}$.

\begin{table*}[t]
\centering
\caption{Overview of the GRW dataset splits. It contains manually verified data for evaluation and finetuning, and large-scale pseudo-labeled data for pre-training.}
\vspace{-10pt}
\label{tab:dataset_splits}
\scriptsize
\setlength{\tabcolsep}{5pt}
\begin{tabular}{l|c|c|c|c|c}
\toprule
\textbf{Task / Split} & \textbf{Total Clips} & \textbf{Hours} & \textbf{Vocab} & \textbf{Semantic} & \textbf{Non-Semantic} \\ 
\midrule

\rowcolor{Gray}
\multicolumn{6}{l}{\textbf{Semantic Gesture Classification}} \\ 
\midrule

Train Split & 135,503 & 149.26 & 155 & 15,340 & 120,163 \\ 
Test Split & 4,000 & 4.41 & 100 & 2,000 & 2,000 \\ 

\midrule
\rowcolor{Blue} 
\textbf{Total} & \textbf{139,503} & \textbf{153.67} & \textbf{155} & \textbf{17,340} & \textbf{122,163}\\

\midrule
\rowcolor{Gray}
\multicolumn{6}{l}{\textbf{Word Recognition \& Localization}} \\ 
\midrule

Pre-train Split (noisy) & 67,214 & 74.26 & 155 & 67,214 & - \\ 
\hline
\hline
Train Split (clean) & 15,340 & 16.94 & 155 & 15,340 & - \\ 
Test Split (clean) & 2,000 & 2.21 & 100 & 2,000 & - \\ 

\midrule
\rowcolor{Blue} 
\textbf{Total (clean)} & \textbf{17,340} & \textbf{19.15} & \textbf{155} & \textbf{17,340} & -\\

\bottomrule
\end{tabular}%
\vspace{-15pt}
\end{table*}

\vspace{-10pt}
\section{Dataset Curation Pipeline}
\label{sec:dataset_pipeline}
\vspace{-10pt}

In this section, we describe the pipeline used to curate the GRW word-level semantic gesture dataset and benchmark. The principal steps are: (i) large-scale automatic mining and filtering from a word-aligned speech corpus; and (ii) multi-stage human annotation with redundancy and verification.  Table~\ref{tab:data_curation} shows the number of clips and the yield of semantic gestures at these stages of the pipeline.

\vspace{5pt}
\subsection{Mining and filtering to obtain candidate video clips}
\vspace{-5pt}

The objective is to obtain a large set of word-aligned video clips potentially containing associated co-speech gestures. 

\begin{table*}[t]
\centering
\caption{\textbf{Overview of the GRW dataset curation pipeline.} We detail the progression of our data from raw video filtering to final manual annotation. After rigorous quality checks, we obtain over 17k annotated semantic clips.}
\vspace{-10pt}
\label{tab:data_curation}
\scriptsize
\renewcommand{\arraystretch}{1.2} 
\resizebox{\textwidth}{!}{%
\begin{tabular}{l|c|c|c|c|c|l}
\toprule
\textbf{Pipeline Stage} & \textbf{Total Clips} & \textbf{Yield} & \textbf{Hours} & \textbf{Vocab} & \textbf{Non-Semantic} & \textbf{Notes} \\ 
\midrule

Raw videos & 1,529,572 & - & 1,699.52 & 200 & n/a & Sourced from MultiVSR\\ 
Mining and filtering & 525,226 & - & 578.40 & 185 & n/a &  Keypoint filtering\\ 
\hline
Assigned for manual labeling & 161,828 & - & 178.21 & 185 & n/a & Output of Section 4.1 \\ 
After recognition annotations & 19,364 & 11.9\% & 21.20 & 155 & 122,163 & Output of Section 4.2 \\ 
After localization annotation & 17,340 & 10.7\% & 19.15 & 155 & n/a & Output of Section 4.3 \\ 

\bottomrule
\end{tabular}%
}
\vspace{-10pt}
\end{table*}

\newpara{Video data source.}
We use the English subset of MultiVSR~\cite{multivsr}, a large-scale multilingual visual speech recognition dataset with word-aligned transcripts, as our in-the-wild video pool. Sourced from YouTube, MultiVSR contains extended full-length versions of the AVSpeech~\cite{ephrat2018looking} videos. The videos exhibit substantial real-world variability across lighting, viewpoints, backgrounds, and speaker demographics. We pre-process the videos to obtain stable upper-body gesture crops. Details on pre-processing are given in the supplementary.

\newpara{Lexicon of gesturable words.} 
Given the MultiVSR English vocabulary, we target \emph{plausibly gesturable words} for iconic, metaphoric, or deictic gestures (e.g., motion verbs, spatial relations, size/shape adjectives), excluding high-frequency function words that rarely elicit semantic depiction. To systematically construct this lexicon, we extract the most frequent lemmas from the dataset's transcripts and prompted an LLM (Gemini~\cite{comanici2025gemini}) to score each word based on its likelihood to elicit a semantic gesture. The top-scoring candidates are then manually verified, yielding a final set of 200 English lemmas expected to induce distinctive gestures. For every match in the dataset to a lemma on this list, we extract a 4-second temporal window centered around the word timestamp. This yields a large collection of candidate clips (1.5M clips), potentially containing semantic gestures. However, this initial pool contains both true positives and substantial noise (e.g., no-gesture segments, beat-only motion, hands out of frame).

\newpara{Visual filtering for gestures.} 
To reduce annotation cost and improve positive sample yield, we apply automatic pose keypoint filtering~\cite{lugaresi2019mediapipe}. For each candidate clip, we compute: (i) the fraction of frames with confident hand/wrist detections, and (ii) the mean temporal hand displacement (velocity), $|p_t - p_{t-1}|_2$, with $p_t$ denoting the frame $t$ hand centroid. Candidates failing minimum visibility or motion thresholds are discarded. This keypoint-based filtering retains approximately 525k candidates across 185 target words. As the mined distribution is highly long-tailed, naive sampling would over-represent high-frequency words. We therefore construct a semi-balanced $\approx$ 162k clip subset for manual annotation, sampling 100 to 1500 instances per word to ensure low-frequency lemma coverage and prevent high-frequency dominance.

\vspace{-10pt}
\subsection{Manual annotation for word recognition}
\label{sec:annotation_word_recog}

\vspace{-5pt}
The objective is to partition the 161,828 candidate clips into two sets: those containing a co-speech gesture corresponding to a target word, and those that don't (e.g., beat or random gestures). Both sets are retained for subsequent model training.

\newpara{Stage-1: Annotating semantic gestures.}
The annotators assign a binary label to each of the candidate samples, indicating the presence of a corresponding semantic gesture for a target word. The annotators are trained to only mark a clip as `Positive' if there is a clearly identifiable gesture that depicts the target word. We developed a custom web-based interface using (VIA)~\cite{dutta2019vgg} (details in the supplementary). For each target word, annotators review a grid of associated video clips (with audio playback) and assign the binary label. This selection stage required approximately 1,000 annotation hours. 

\newpara{Stage-1: Quality control.} Each sample is independently labeled by five annotators, screened using a held-out set of 500 gesture clips. To reduce mistakes post-annotation, we aggregate labels via majority voting, discarding words with high disagreement, yielding roughly 19k semantic videos (11.9\% positive rate). 

\vspace{-10pt}
\subsection{Manual annotation for word localization}

\vspace{-5pt}
\newpara{Stage-2: Annotating temporal gesture boundaries.} The input to this stage is the set of 19k output semantic gesture clips obtained from Section~\ref{sec:annotation_word_recog}. For each clip, annotators mark the start-end timestamps of the gesture corresponding to the target word. This task is more demanding than binary recognition, as it requires precise temporal localization through scrubbing, frame stepping, and repeated viewing. The interface and detailed annotation guidelines are provided in the supplementary. This stage required $\approx$ 500 hours of annotation efforts.

\newpara{Stage-2: Quality control.} To ensure gesture boundary reliability, we retain only those samples for which \textit{at least three annotators} provide start-end timestamps that differ by less than 1-second. This filtering yields 17,340 samples containing semantically grounded gestures with reliable word-level temporal boundaries, which becomes our final GRW semantic dataset.  This amounts to 19.15 hours of clean data. 

\vspace{-10pt}
\subsection{Curating a pre-train set}
\label{subsec:pretrain}

\vspace{-5pt}
To scale downstream word recognition and temporal localization without prohibitive manual annotation costs, we employ a pseudo-labeling strategy. By applying a strict confidence threshold to the classifier's predicted probability of a semantic gesture ($\ge 0.9$) to our trained semantic classifier (Section~\ref{sec:semantic_classification_model}), we automatically mine 67,214 high-confidence positive clips from the $\approx 363$k unannotated candidates (Table~\ref{tab:data_curation}). For weak localization labels, we pad each clip's speech boundaries using word-specific average start-end frame offsets derived from the clean GRW train split. This weakly-labeled set is then used to pre-train our recognition and localization models.

\vspace{-10pt}
\section{Dataset Analysis}
\label{sec:dataset_analysis}

\vspace{-10pt}
In this section, we present an initial analysis of the co-speech gestures in the GRW dataset. We are interested in three questions: (i) how likely a given word is to be accompanied by a semantically meaningful gesture; (ii) how the timing of the gesture segment relates to the spoken word segment; and (iii) how variable the gestures are for the same word across speakers and context. Prior works~\cite{liu2022beat,hegde2025understanding} have investigated some of these questions, but have been in laboratory settings or with limited scale and diversity of the data -- potentially restricting the scope and generality of their findings.

\vspace{-10pt}
\subsection{How likely are the words to be gestured?}
\vspace{-5pt}

\begin{figure}
    \centering
    \includegraphics[width=\linewidth]{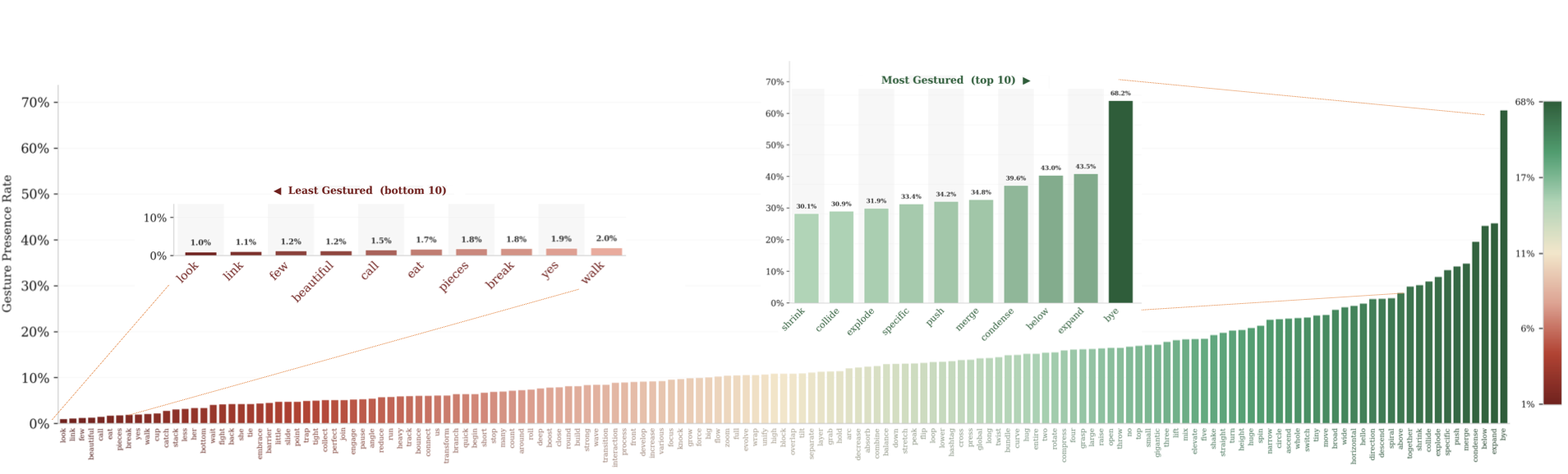}
    \vspace{-15pt}
    \caption{\textbf{Not all words are gestured equally}: We plot the likelihood that a spoken word is accompanied by a depictive semantic gesture. Abstract concepts and common verbs (left inset) rarely elicit semantic gestures ($<2\%$), whereas inherently spatial or physical actions like push, below, and bye (right inset) are gestured with high frequency.}
    \label{fig:gestured_words}
    \vspace{-15pt}
\end{figure}

Fig~\ref{fig:gestured_words} shows the proportion of gestured instances among all clips submitted for annotation. The likelihood that a word is accompanied by a semantically meaningful gesture varies dramatically across the lexicon. While some words are gestured very frequently, for example, `bye' is gestured approximately 68\%, while others are rarely depicted. At the extreme low end, `look' is gestured in 1\% of cases. This long-tailed distribution highlights the sparsity of semantic gestures in natural discourse and motivates the need for large-scale filtering and careful annotation to obtain high-quality gesture–word pairs. 

\vspace{-20pt}

\vspace{-10pt}
\subsection{Gesture vs.\ speech boundaries}
\vspace{-5pt}

\begin{figure}
    \centering
    \includegraphics[width=\linewidth]{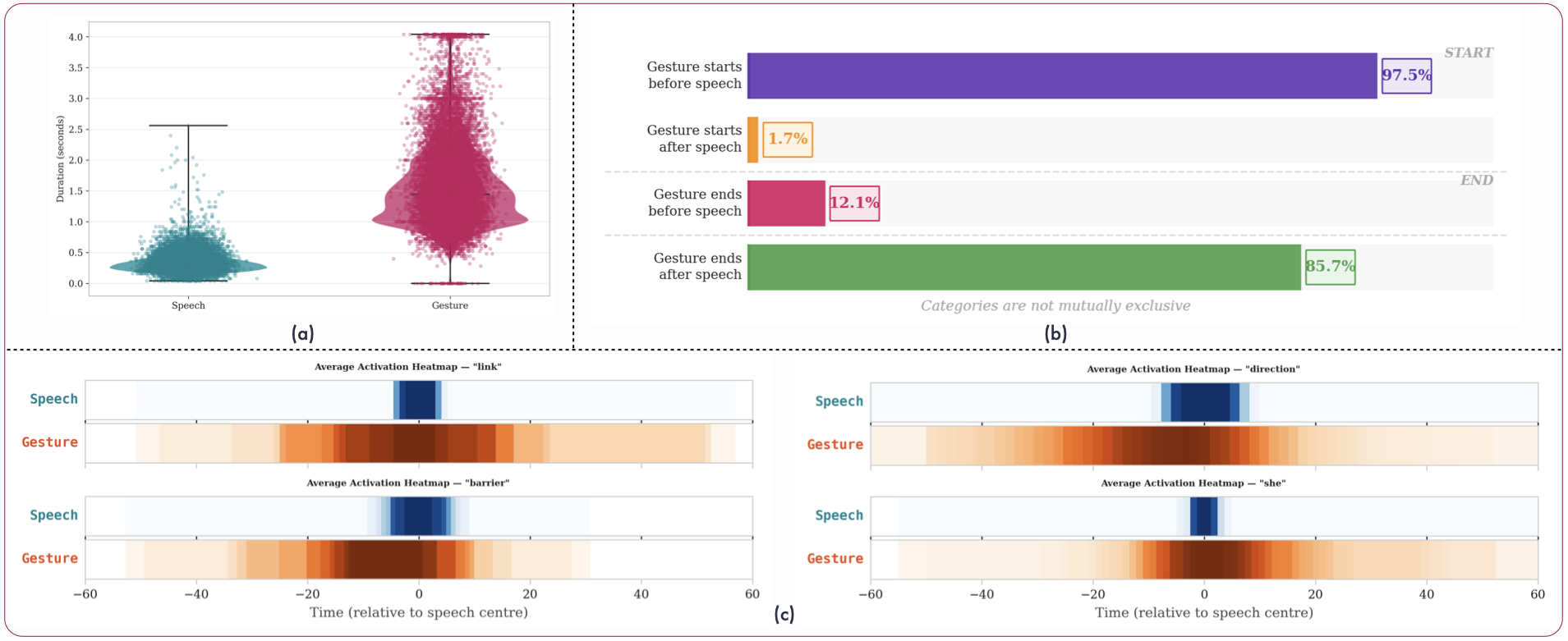}
    \vspace{-15pt}
    \caption{\textbf{Gesture-speech temporal alignment}. Semantic gestures rarely align perfectly with spoken words. As shown in (a), gestures are significantly longer in duration than speech. (b) The vast majority of gestures start before (97.5\%) and end after (85.7\%) the target word is spoken. (c) Activation heatmaps (aggregated across all samples for a specific word) visually confirm this ``envelope'' effect.}
    \label{fig:alignment_stats}
    \vspace{-5pt}
\end{figure}

We analyze the temporal relationship between spoken words and their semantic gestures on a larger scale and find that gesture–speech alignment is loose and highly variable. The Violin plot in 
Fig~\ref{fig:alignment_stats}(a) shows that the average duration of spoken word segments is much shorter than their corresponding gesture segments across all annotated words.  
Fig~\ref{fig:alignment_stats} (b) quantifies this mismatch. Gestures start before spoken word (97.5\%) and end after spoken word (85.7\%). This also corroborates the findings in~\cite{ter2025co, ghaleb2024leveraging}. Finally, Fig~\ref{fig:alignment_stats} (c) visualizes averaged activation heatmaps for a few word classes, showing that gestures are often distributed asymmetrically around the speech center. 
Table~\ref{tab:durations} compares the average durations of speech and gestures for selected words. Together, these analyses highlight that gesture boundaries are loosely aligned with speech, making word-level gesture recognition and gesture localization particularly challenging.

\vspace{-5pt}
\begin{table}
  \caption{Average speech and gesture durations for a few words from the GRW dataset. Values are (speech | gesture) in seconds.}
  \vspace{-10pt}
  \label{tab:durations}
  \centering
  \scriptsize
  \setlength{\tabcolsep}{4pt}
  \begin{tabular}{c|c||c|c||c|c||c|c||c|c||c|c||c|c}
    \toprule 
    \multicolumn{2}{c||}{\textbf{join}} & \multicolumn{2}{c||}{\textbf{condense}} & \multicolumn{2}{c||}{\textbf{two}} & \multicolumn{2}{c||}{\textbf{interaction}} & \multicolumn{2}{c||}{\textbf{knock}} & \multicolumn{2}{c||}{\textbf{spiral}} & \multicolumn{2}{c}{\textbf{hello}}\\ 

    \midrule

    0.31 & 1.81 & 0.50 & 1.30 & 0.22 & 1.34 & 0.60 & 2.07 & 0.30 & 1.56 & 0.46 & 2.17 & 0.34 & 1.01\\

    \bottomrule
  \end{tabular}%
  \vspace{-15pt}
\end{table}

\vspace{-10pt}
\subsection{Variations in gestures}
\vspace{-5pt}

Co-speech gestures vary substantially across speakers and contexts: a single word can elicit multiple distinct movements, and different words may share visually similar gestures. For example (Fig.~\ref{fig:dataset}), speakers depict `spiral' using one or both hands, via vertical or horizontal motions. Similarly, `zoom' elicits spreading palms, opening fingers, or moving hands apart. Conversely, `shrink' is gestured by bringing hands inward, pinching fingers, or pressing palms together. 

\vspace{-10pt}
\section{Recognizing Co-speech Gestures}
\label{sec:method}

\begin{figure}
    \centering
    \includegraphics[width=\linewidth]{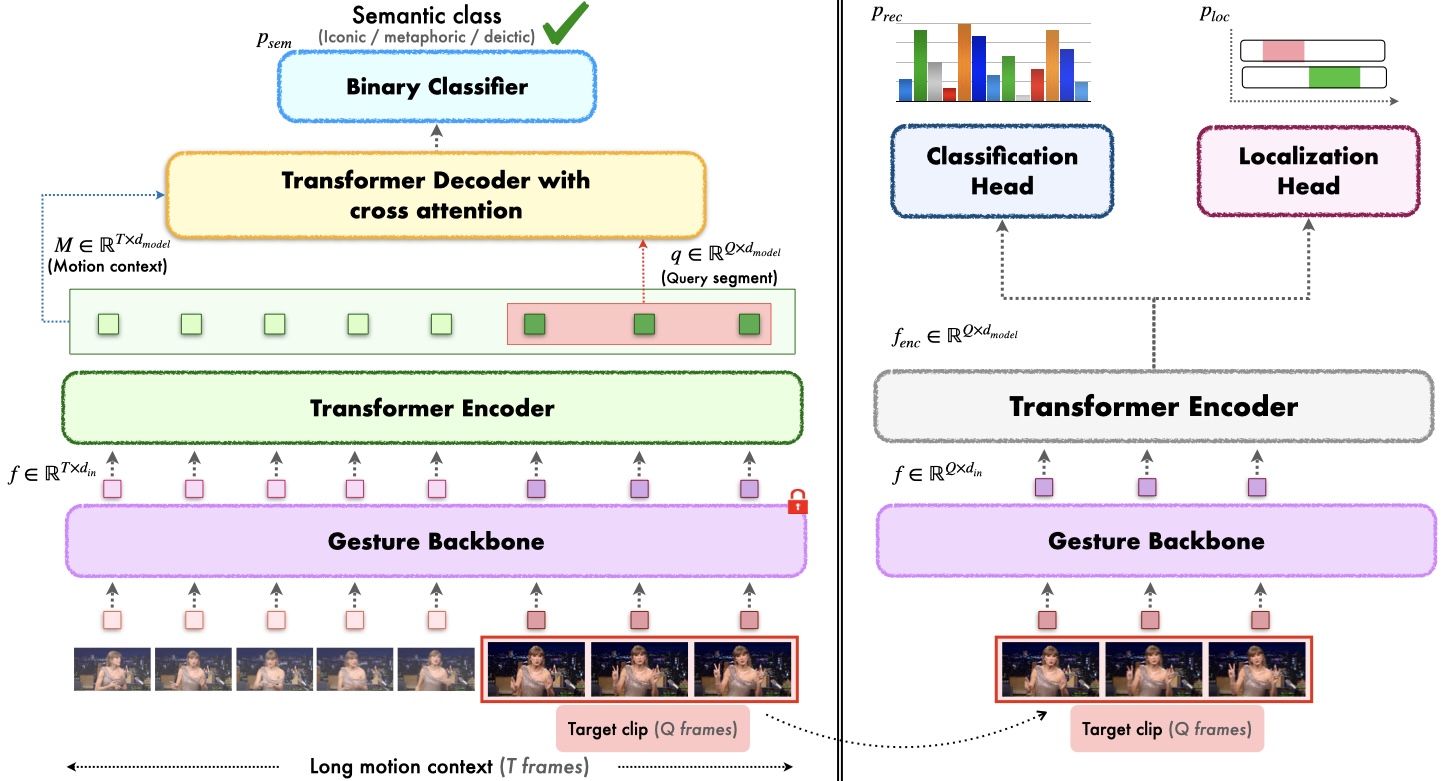}
    \vspace{-15pt}
    \caption{\textbf{Left:} A binary semantic gesture classifier for a query clip, given a broad temporal motion context. The model uses cross-attention between a target query clip and the motion context to determine if the gesture within the query clip is semantic or not. \textbf{Right:} The word recognition and localization model operates on the query clip and is trained to classify the word class, and the precise temporal boundary of the gesture (per-frame binary labels).}
    \vspace{-20pt}
    \label{fig:architecture}
\end{figure}

\vspace{-5pt}
We introduce two complementary co-speech gesture recognition models that are trained on GRW: (i) a \textbf{semantic gesture classification} model that predicts whether a short video clip contains a clear semantic (iconic, deictic, metaphoric) gesture or not, and (ii) a \textbf{gestured word recognition \& localization} model that (a) predicts the gestured word class, and (b) localizes the gesture segment in time within a short clip. The architecture of the two models is illustrated in Fig~\ref{fig:architecture}. Both models are built on top of a frozen visual backbone pretrained to extract per-frame hand shape and position features. The features are then ingested by a series of transformer blocks for task-specific temporal modeling and prediction. In the following, we first describe the common visual backbone and then the two task specific transformer architectures.

\vspace{-10pt}

\vspace{-3pt}
\subsection{Gesture backbone}
\label{backbone}

\vspace{-3pt}
Given a video clip $V \in \mathbb{R}^{T \times H \times W \times 3}$, we use SHuBERT~\cite{gueuwou2025shubert} as our pretrained gesture backbone to extract frame-level features. Pretrained on large-scale sign-language data~\cite{uthus2023youtube}, SHuBERT provides robust visual features that inherently capture fine-grained hand shapes and semantically distinctive motions. We extract the hidden states from $L$ layers of this Transformer-based model to construct a multi-layer sequence tensor $f_{\text{multi}} \in \mathbb{R}^{L \times T \times d_{in}}$. Here, $d_{in} = 768$ represents the feature dimension, and $T$ is the number of frames (corresponding to approximately 4-10 seconds of video).

\newpara{Weighted layerwise feature aggregation.} To optimally fuse SHuBERT's layers, an MLP processes $f_{multi}$ to generate $L$ scalar scores, which are softmax-normalized into mixing weights $\alpha$. A weighted sum across the layer dimension then collapses this into a single sequence $f \in \mathbb{R}^{T \times d_{in}}$, serving as the direct input to our downstream task-specific models (see supplementary for details).

\vspace{-12pt}
\subsection{Semantic Co-speech Gesture Classification}
\label{sec:semantic_classification_model}
\vspace{-5pt}
The semantic gesture classifier is designed to answer: \emph{Does a  video clip contain a clear semantic gesture or not, given the surrounding motion context?} Note that the ``motion-context'' is important here to spot semantic gestures in a reliable way in unconstrained co-speech gestures. 

\newpara{Intuition.} Semantic gestures are typically sparse, arbitrary, and highly variable across different words and are not necessarily unique to a person. In contrast, non-semantic beat gestures are frequent, repetitive, and serve as a person's default gesturing rhythm. By encoding a long temporal window, the model learns the speaker's baseline kinematics for non-semantic beat gestures and for their random gestures. This context makes it significantly easier for the model to contrast, isolate, and identify the sparse, unique semantic gestures, as such gestures physically deviate from the speaker's established baseline.

\newpara{Task formulation.} Given a \emph{long} context video with $T$ frames and a \emph{query interval} $q = [s,e]$ with $Q=e-s+1$, the goal is to predict a binary label $y \in \{0,1\}$, where $y=1$ indicates that the query interval contains a \emph{clear semantic gesture} and $y=0$ indicates no semantic gesture. 

\newpara{Architecture.} Visual features $f$ are projected to $d_{model}$ via a 2-layer MLP (LayerNorm, ReLU) and processed by a Transformer Encoder. To model relative motion timing across the long context (up to 10 seconds), we apply Rotary Position Embeddings (RoPE)~\cite{rotary}, yielding a context-aware memory $M \in \mathbb{R}^{T \times d_{model}}$. Features spanning the target interval $[q_{start}, q_{end}]$ are extracted to form a query sequence $q \in \mathbb{R}^{Q \times d_{model}}$. This query cross-attends to the full memory $M$ within a Transformer Decoder, effectively referencing the full motion context when interpreting the query interval. Finally, the decoder output is temporally mean-pooled and passed through an MLP with a sigmoid activation to yield the binary semantic probability $p_{sem}$.

\newpara{Training.} We train the semantic gesture classifier using the `Semantic Gesture Classification' split of the GRW dataset in Table~\ref{tab:dataset_splits}, consisting of 135,503 training examples, which exhibits a natural class imbalance (15,340 positive semantic gesture samples and 120,163 negative samples). To address this, we use class-balanced sampling to upsample positive class data in each batch, and additionally apply a class-weighting to the loss (assigns higher weight to positive samples). During training, we extract a long temporal context of $T=250$ frames (10 seconds) and define a target query interval of $Q=100$ frames (4 seconds). The model is optimized using a Binary Cross-Entropy (BCE) loss.

\vspace{-10pt}
\subsection{Co-speech Gesture Word Recognition and Localization}

\vspace{-4pt}
Given a \emph{short} query clip containing a semantic gesture somewhere inside it, the goal is to predict: (i) the class of the word: $c \in \{1,\ldots,C\}$ corresponding to the co-speech gesture, and (ii) per-frame gesture presence: $\mathbf{a} = [a_1,\ldots,a_Q],\;\; a_q \in \{0,1\}$, where $a_q=1$ indicates that frame $q$ lies within the gesture interval and $a_q=0$ otherwise. We use the SHuBERT features $f_{multi}$, which are then aggregated layer-wise to obtain $f \in \mathbb{R}^{Q \times d_{in}}$, same as in Section~\ref{backbone}. These features are then temporally encoded using a Transformer encoder to get $f_{enc}$.  

\newpara{Gesture localization.} To achieve precise temporal localization without requiring complex region-proposal pipelines, we formulate the alignment task as a dense, frame-level binary classification. An MLP + sigmoid operates directly on per-frame encoder features $f_{enc}$ to get per-frame probabilities ${p_{loc}}$.

\newpara{Word classification.} We do a weighted average of the Transformer output features $f_{enc}$ across $Q$ frames to form a single, global video representation. The weights are determined by $p_{loc}$, which provides a per-frame probability of the relevant gesture being  present. A classification head consisting of a two-layer MLP projects this global embedding to produce softmax probabilities $p_{rec}$ over the $C$ semantic word classes.

\vspace{-25pt}
\subsection{Training the Word Recognition and Localization Model} 

\vspace{-5pt}
The training is divided into two steps: (i) the model is pre-trained on the pseudo-labeled GRW pre-training data (67k samples) -- this is weak supervision because the pseudo-labeled data only has approximate gesture boundaries (only the word is known); (ii) the model is fine-tuned on the manually annotated GRW word recognition training dataset (15k samples) -- this is strong supervision as the gesture boundaries are precisely annotated.

\newpara{Weakly-supervised pre-training.} We pre-train the word recognition and localization model on the 67,214 pseudo-labeled positive samples. Frames falling within the padded speech boundaries are treated as positives, and the rest are negative frames. The model is supervised using frame-level BCE loss for temporal localization, combined with a categorical cross-entropy loss for gestured word classification: 
\setlength{\abovedisplayskip}{-1pt}
\setlength{\belowdisplayskip}{-1pt}
$$\mathcal{L} = - \sum_{c=1}^{C} y_{c} \log(p_{rec, c}) - \lambda_{loc} \frac{1}{Q} \sum_{q=1}^{Q} \left[ a_q \log(p_{loc, q}) + (1-a_q) \log(1-p_{loc, q}) \right]$$
where $y_c$ is the one-hot word label, $p_{rec, c}$ is the predicted word class probability, $a_q$ is the `ground-truth' binary frame label, and $p_{loc, q}$ is the predicted frame-level gesture probability at frame $q$.

\newpara{Fine-tuning on clean data.} Finally, we fine-tune the recognition and localization model exclusively on the manually verified train split of 15,340 samples (Table~\ref{tab:dataset_splits}). We use the same loss functions as in the pre-training stage. 

\vspace{-10pt}
\section{Results}
\label{sec:results}
\vspace{-5pt}

\begin{table}[tb]
  \caption{\textbf{Semantic gesture classification performance}. Our approach surpasses existing baselines, especially at high confidence thresholds ($\ge 0.9$). This validates its utility for extracting clean semantic gestures from unlabeled data.}
  \vspace{-10pt}
  \label{tab:results_semantic}
  \centering
  \scriptsize
  \renewcommand{\arraystretch}{1.2}
  \resizebox{\textwidth}{!}{%
  \begin{tabular}{l|c|c|c|c}
    \toprule 
    \textbf{Method} & \textbf{Accuracy (\%)} & \textbf{Precision (\%)} & \textbf{Recall (\%)} & \begin{tabular}{@{}c@{}}\textbf{Semantic class accuracy (\%)} \\ \textbf{for high-conf samples}\end{tabular} \\
    \midrule

    Random & 50.00 & 50.00 & 50.00 & n/a\\
    \hline
    \rowcolor{Gray} 
    \multicolumn{5}{l}{\textbf{Frozen features}} \\
    \hline
    Clip4Clip~\cite{luo2022clip4clip} & 59.13 & 55.64 & 52.34 & 62.95 \\
    Sapiens~\cite{khirodkar2024sapiens} & 60.10 & 60.92 & 57.80 & 68.01\\
    SemMom$_{Dinov3}$~\cite{semanticmoments2026} & 61.77 & 60.91 & 58.85 & 67.31\\
    \hline
    
    \rowcolor{Gray} 
    \multicolumn{5}{l}{\textbf{Fine-tuned features}} \\
    \hline
    GestSync~\cite{hegde2023gestsync} & 61.58 & 59.90 & 62.17 & 64.91 \\
    JEGAL~\cite{hegde2025understanding} & 63.94 & 61.02 & 59.94 & 66.93 \\
    \hline
    
    \rowcolor{Gray} 
    \multicolumn{5}{l}{\textbf{VLMs}} \\
    \hline
    Intern-VL~\cite{chen2024internvl} & 61.58 & 56.67 & 58.35 & 66.68\\
    Qwen3-VL~\cite{bai2025qwen3} & 55.05 & 53.24 & 51.28 & 58.61\\
    Gemini-3.5~\cite{comanici2025gemini} & \underline{67.30} & \underline{65.78} & \textbf{71.10} & \underline{68.21} \\
    
    \midrule
    \rowcolor{Blue}
    \textbf{Ours} & \textbf{75.83} & \textbf{79.91} & \underline{69.00} & \textbf{93.20} \\
    \bottomrule
    
  \end{tabular}
  }
  \vspace{-18pt}
\end{table}

\vspace{-5pt}
In this section, we evaluate on the unseen test set of GRW for semantic gesture classification  (Section~\ref{sec:classification_results}), and gestured word recognition \& localization (Section~\ref{sec:loc_results}). In addition to evaluating on the test sets of GRW, we also evaluate the gesture classifier on unseen words, and the localization ability of the gesture recognizer on the test set of~\cite{hegde2025understanding}. The baseline models used for comparison and the task-specific evaluation metrics are described in the corresponding subsections.

\vspace{-15pt}
\subsubsection{Baselines.}
We evaluate on a diverse set of baselines, starting with strong pretrained video-language representations as \emph{frozen} feature extractors, training only a lightweight classifier on top. We use Clip4Clip~\cite{luo2022clip4clip}, Sapiens~\cite{khirodkar2024sapiens}, and SemanticMoments (on Dinov3 features)~\cite{semanticmoments2026}, to measure how far generic pretrained representations go \emph{without} end-to-end adaptation. Next, we compare with the recent state-of-the-art co-speech gesture models that learn dense frame-level representations. We finetune GestSync~\cite{hegde2023gestsync} and JEGAL~\cite{hegde2025understanding} with additional task-specific prediction heads on the exact same data as our approach. Finally, we evaluate two state-of-the-art open-source VLMs, InternVL~\cite{chen2024internvl}, Qwen3-VL~\cite{bai2025qwen3} and a production-grade VLM, Gemini-3.5 in a prompted setting: given a video clip and a list of candidate word classes, the model is asked to: (i) classify if the video clip contains a semantic gesture, and (ii) recognize the gestured word and localize the temporal boundaries (start and end time) of the corresponding gesture instance. Detailed prompts are provided in the supplementary material.

\vspace{-15pt}
\subsection{Semantic Gesture Classification}
\label{sec:classification_results}

\vspace{-5pt}
\noindent
\textbf{Metrics:}
We evaluate performance using \textit{Accuracy}, \textit{Precision}, and \textit{Recall} to account for inherent class imbalance. Additionally, to evaluate the classifier's reliability for generating pseudo-labels, we report its accuracy specifically on high-confidence predictions ($p_{sem} \ge 0.9$).

\noindent
\textbf{Results:} 
Table~\ref{tab:results_semantic} demonstrates that our model outperforms baseline methods across the board. Notably, it excels at high-confidence prediction (accuracy of 93.2\%), confirming that utilizing a $\ge 0.9$ threshold effectively isolates semantic gestures in unlabeled data with minimal label noise, although this comes at the cost of lower recall (35\%).

\newpara{Is the semantic classifier word-class agnostic?} We evaluate this aspect by creating a new test set with 10 unseen words. Specifically, we manually annotate 500 new clips with binary labels indicating whether the clips contain a semantic gesture or not, with 201 clips containing semantic gestures. We ensure that these clips do not contain any words that are present in the GRW dataset. We measure semantic class accuracy at high confidence thresholds for both our model and Gemini, which comes out to be 86.3\% and 63.8\% respectively, with a recall of 31.3\% and 35.2\% respectively. This indicates that our model generalizes across word classes and can be used for large-scale semantic gesture mining in future work. 

\vspace{-14pt}
\subsection{Word Recognition and Localization}
\label{sec:loc_results}

\vspace{-5pt}
\noindent
\textbf{Metrics:} We report top-k accuracy for recognition on the unseen test set of 100 word classes, for $k \in \{1, 5, 10\}$. For localization, we report mIoU to measure the temporal overlap between the ground-truth and predicted gesture segment.

\noindent
\textbf{Results:}
Table~\ref{tab:results_recog} shows the word recognition and localization results. We clearly outperform all the baselines on both tasks by a large margin. Gemini is particularly weaker on fine-grained gesture localization task. We also note that Acc@5 is a more indicative measure of performance due to the ambiguity between certain classes; for example, pairs such as \emph{(bye, hello)} and \emph{(little, small)} are often expressed using similar gestures. 
\begin{table}[tb]
  \caption{Word-level gesture recognition \& localization performance. Our model outperforms baselines on top-$k$ recognition accuracy and temporal localization (mIoU).}
  \vspace{-10pt}
  \label{tab:results_recog}
  \centering
  \scriptsize
  \setlength{\tabcolsep}{10pt}
  \begin{tabular}{l|c|c|c||c}
    \toprule 
    \multirow{2}{*}{\textbf{Method}} & \multicolumn{3}{c||}{\textbf{Recognition}} & \textbf{Localization} \\ 
    \cline{2-5}
    & \textbf{Acc @ 1} & \textbf{Acc @ 5} & \textbf{Acc @ 10} & \textbf{mIoU} \\
    \midrule

    Random & 1.00 & 5.00 & 10.00 & 0.1975\\
    
    \hline
    \rowcolor{Gray} 
    \multicolumn{5}{l}{\textbf{Frozen features}} \\
    \hline

    Clip4Clip~\cite{luo2022clip4clip} & 8.70 & 19.25 & 27.60 & n/a\\
    Sapiens~\cite{khirodkar2024sapiens} & 9.45 & 18.80 & 28.60 & n/a\\
    SemMom$_{Dinov3}$~\cite{semanticmoments2026} & 10.05 & 21.15 & 29.65 & n/a \\

    \hline
    \rowcolor{Gray} 
    \multicolumn{5}{l}{\textbf{Fine-tuned features}} \\
    \hline

    GestSync~\cite{hegde2023gestsync} & 9.12 & 20.37 & 30.18 & 0.5172 \\
    JEGAL~\cite{hegde2025understanding} & 10.43 & 22.71 & 31.88 & \underline{0.5368}\\

    \hline
    \rowcolor{Gray} 
    \multicolumn{5}{l}{\textbf{VLMs}} \\
    \hline

    Intern-VL~\cite{chen2024internvl} & 4.35 & 12.50 & 15.71 & 0.3932\\
    Qwen3-VL~\cite{bai2025qwen3} & 5.92 & 13.19 & 19.27 & 0.2071\\
    Gemini-3.5~\cite{comanici2025gemini} & \underline{13.95} & \underline{30.15} & \underline{41.10} & 0.4658\\

    \midrule
    \rowcolor{Blue}
    \textbf{Ours} & \textbf{18.35} & \textbf{37.30} & \textbf{51.70} & \textbf{0.6731} \\
    
  \bottomrule
  \end{tabular}%
  \vspace{-10pt}
\end{table}

\vspace{-10pt}
\begin{figure}[ht]
    \centering
    \includegraphics[width=\linewidth]{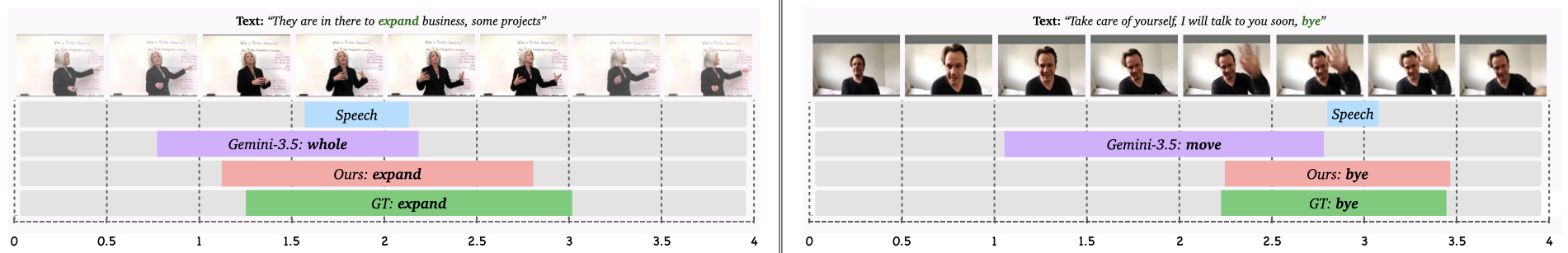}
    \vspace{-20pt}
    \caption{We visualize sample predictions for joint word recognition and localization. In both instances, our model correctly assigns the target word (expand, bye) and closely matches the ground truth temporal gesture boundaries.} 
    \label{fig:qualitative_results}
    \vspace{-20pt}
\end{figure}

\begin{figure}
    \centering
    \includegraphics[width=4.6in, height=1.8in]{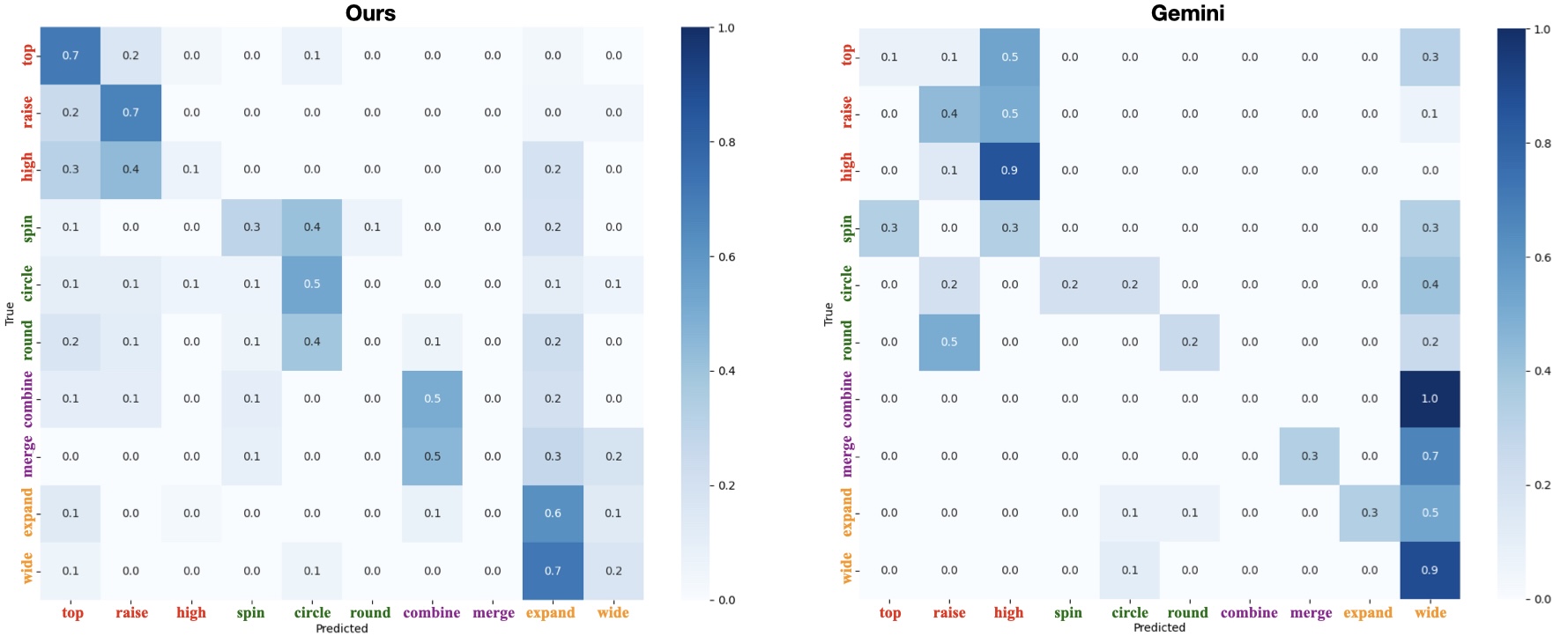}
    \vspace{-5pt}
    \caption{Comparison of the confusion matrices for our model and Gemini. Most misclassifications occur between semantically related words.}
    \label{fig:confusion_matrix}
    \vspace{-3pt}
\end{figure}

Fig~\ref{fig:qualitative_results} provides qualitative examples demonstrating the model's accurate word recognition together with precise gesture localization results. Additionally, the confusion matrices in Fig.~\ref{fig:confusion_matrix} show that most misclassifications occur between semantically related words, highlighting the inherent ambiguity in gesture-based word recognition.

\newpara{Generalization to other datasets.} We evaluate on the AVS-Spot benchmark~\cite{hegde2025understanding}, which contains 500 samples with approximate temporal boundary annotations. The goal is to localize the semantic gesture in time, and the prediction is treated as correct if the predicted gesture segment has any overlap with the ground-truth segment. We obtain the gesture timestamps from our model's localization head, and compare the semantic gesture spotting accuracy directly with JEGAL~\cite{hegde2025understanding}. We achieve an accuracy of \textbf{77.3\%} compared to JEGAL's 63.6\%, which is a significant improvement on an existing gesture benchmark.

\vspace{-10pt}

\subsection{Ablation Studies}
\vspace{-10pt}

\newpara{Impact of context length.} For the semantic gesture classification task, our approach leverages a substantially longer temporal context to determine the presence of a semantic gesture within a short target clip. As shown in Table~\ref{tab:ablation_context}, providing 10 seconds of temporal context for classifying a 4 second target segment improves accuracy by 5.9\%.

\vspace{-20pt}
\begin{table}
  \caption{Providing a temporal context of 10s (250 frames) to make a prediction for a 4s (100 frames) target segment improves the accuracy by 5.9\%.}
  \vspace{-10pt}
  \label{tab:ablation_context}
  \centering
  \scriptsize
  \begin{tabular}{l|c|c|c|c|c}
    \toprule 
    & \textbf{100 frames} & \textbf{150 frames} & \textbf{200 frames} & \textbf{250 frames} & \textbf{300 frames} \\
    \midrule

    \textbf{High-conf Acc. (\%)} & 87.27 & 89.82 & 91.56 & \textbf{93.20} & 92.95 \\

    \bottomrule
  \end{tabular}
  \vspace{-10pt}
\end{table}

\vspace{-2pt}
\begin{table}
\vspace{-8pt}
  \caption{Pre-training before fine-tuning on the clean train set gives a clear performance improvement across all the metrics.}
  \vspace{-10pt}
  \label{tab:ablation_pretraining}
  \centering
  \scriptsize
  \setlength{\tabcolsep}{4pt}
  \begin{tabular}{l|c|c|c||c}
    \toprule 
    \multirow{2}{*}{\textbf{Method}} & \multicolumn{3}{c||}{\textbf{Recognition}} & \textbf{Localization} \\ 
    \cline{2-5}
    & \textbf{Acc @ 1} & \textbf{Acc @ 5} & \textbf{Acc @ 10} & \textbf{mIoU} \\
    \midrule

    \textbf{Train using Pre-train} & 11.50 & 28.15 & 38.70 & 0.5737 \\
    \textbf{Train using Clean} & 16.35 & 35.60 & 49.40 & 0.6473\\

    \rowcolor{Blue}
    \textbf{Pre-train + FT on clean} & \textbf{18.35} & \textbf{37.30} & \textbf{51.70} & \textbf{0.6731}\\
    
  \bottomrule
  \end{tabular}%
  \vspace{-15pt}
\end{table}

\newpara{Impact of pre-training.}
Table~\ref{tab:ablation_pretraining} shows that pre-training on noisy pre-training set improves both recognition and localization performance. Using manually verified labels provides a further boost. Interestingly, even without any manually labeled data, the model achieves competitive performance, suggesting that large-scale weakly supervised pre-training is a promising approach for these tasks.

\vspace{-12pt}
\section{Conclusion}
\label{sec:conclusion}

\vspace{-8pt}
We have introduced the GRW dataset, providing a large-scale benchmark for analysing, training, and evaluating  models for in-the-wild co-speech gestures. We study the temporal offsets between speech and gestures, and also propose new models for three downstream gesture tasks using the dataset. We acknowledge, however, that our findings are conditioned on the nature of our source material -- since GRW is derived from public-facing discourse such as lectures, talk shows, and interviews, the observed behaviors may reflect a specific register of formal communication. The gesture patterns may be slightly different in social settings or private conversations~\cite{bavelas1995gestures}. Despite this caveat, GRW represents a significant leap in scale and annotation granularity over existing benchmarks, enabling models to associate semantic gestures with spoken words.

\vspace{2mm}

\footnotesize{
\noindent \textbf{Acknowledgements.}
We would like to thank Taein Kwon, Piyush Bagad, Jaesung Huh, and Paul Engstler for  valuable discussions. We are grateful to Alyosha Efros, Jitendra Malik, and Justine Cassell for their valuable feedback and insightful suggestions; and to Abhishek Dutta, Prasanna Sridhar and David Pinto for setting up the data annotation tool, and  Ashish Thandavan for infrastructure support. We also thank Elancer IT Solutions for their assistance with data annotation. This research is funded by EPSRC Programme Grant VisualAI EP/T028572/1 and a Royal Society Research Professorship RSRP\textbackslash R\textbackslash 241003. Sindhu is supported by a Google PhD Fellowship.
}

\bibliographystyle{splncs04}
\bibliography{main}

\newpage
\clearpage
\appendix

\section{Dataset}
\subsection{Data samples}

\begin{figure}
    \centering
    \includegraphics[width=\linewidth]{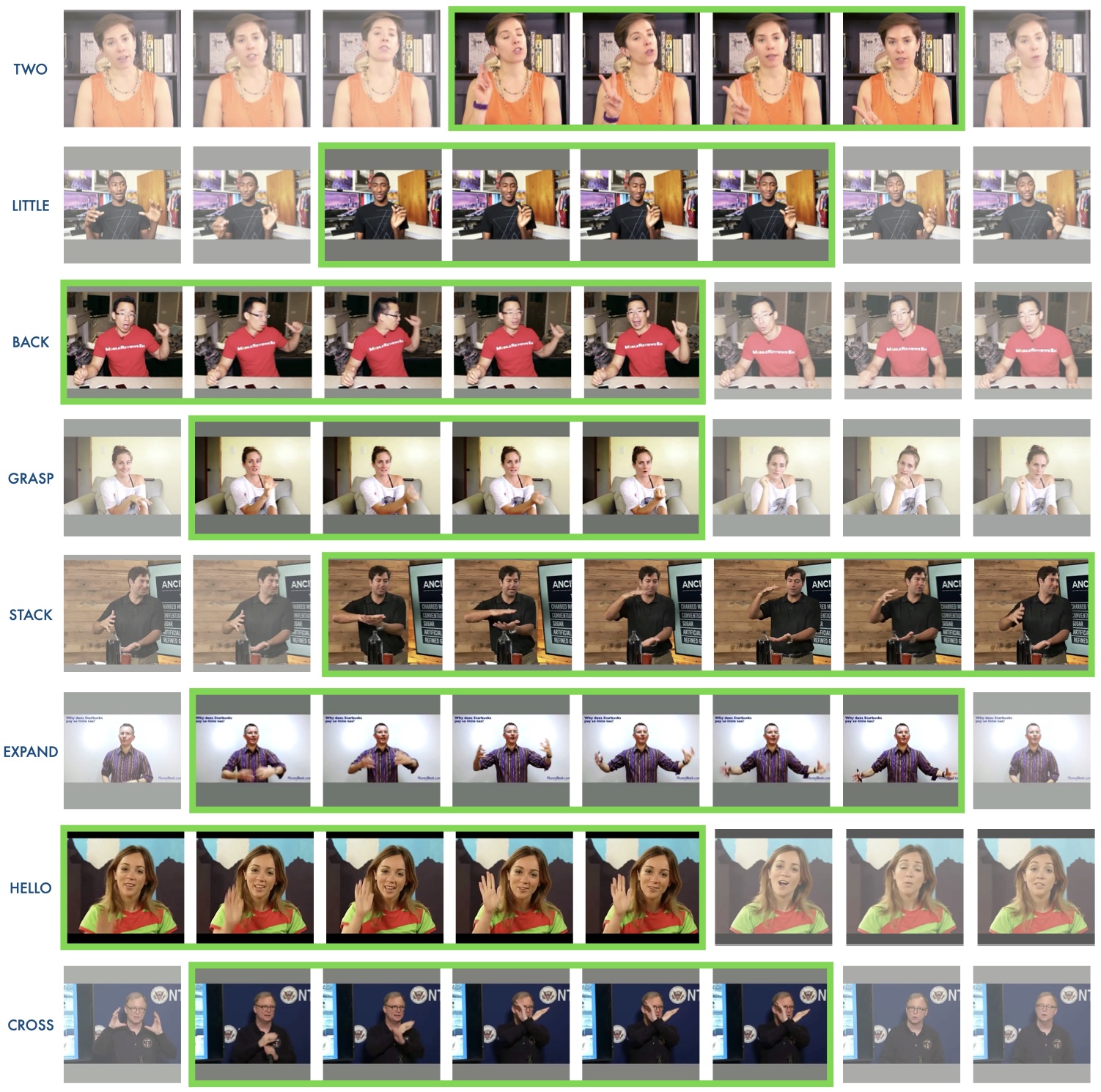}
    \caption{\textbf{Data samples from the GRW dataset.} The dataset encompasses a diverse vocabulary of 155 conceptual words, ranging from numbers (`two') and spatial descriptors (`little', `back', `expand', `cross') to physical actions (`grasp', `stack') and conversational anchors (`hello'). For each positive instance, we provide precise, frame-accurate temporal boundaries marking the start and end of the semantic gesture (highlighted by the green outlines), distinguishing the meaning-bearing motion from surrounding frames.}
    \label{fig:data_samples}
\end{figure}

Fig~\ref{fig:data_samples} illustrates representative samples from the semantic subset of our manually annotated GRW dataset. The dataset comprises 155 unique words, carefully curated to capture a diverse range of clear, well-defined, and semantically meaningful gestures. Video samples can be found in the supplementary video.

\subsection{Dataset bias}
Our dataset is derived from MultiVSR~\cite{multivsr}, which consists primarily of celebrity interviews, television shows, lectures, and conversational videos. While this source provides coverage across a wide range of topics and speaking styles, it may nevertheless introduce inherent biases. For instance, the distribution of speakers, settings, and interaction contexts may not be representative of the broader population. Gestural behavior in media settings, such as interviews or staged discussions, can differ from spontaneous, everyday communication, potentially influencing the types and frequency of gestures captured in the dataset. Moreover, all videos in GRW are in English. As a result, the dataset predominantly reflects gesture patterns of English-speaking individuals, many of whom may share similar cultural norms. These factors should be considered when interpreting results and analysis made on the GRW dataset.

\section{Annotation}
\label{sec:dataset_pipeline}

We provide additional details on the annotation process below.

\subsection{Obtaining candidate words}
The following is the prompt provided to Gemini, together with a list of highly frequent words in the dataset that need to be scored on how likely they are to induce a semantic gesture.

\begin{tcolorbox}[fontupper=\scriptsize, title=Prompt to obtain our dataset lexicon, colback=gray!5!white, colframe=gray!75!black]
\textbf{Role:} You are an expert linguist and researcher specializing in human co-speech gesture understanding and multimodal communication.

\textbf{Task:} You will be provided with a plain text file containing a list of English lemmas (one per line). Your objective is to evaluate each word's ``gesturability'': the likelihood that a speaker will naturally induce a visually depictive, semantic co-speech gesture when uttering this word in unconstrained, natural discourse.\\

\textbf{Definitions to apply:}
\begin{itemize}
    \item \textbf{Iconic:} Gestures that physically depict the shape, size, or action of the target word (e.g., drawing a circle in the air for ``round'').
    \item \textbf{Metaphoric:} Gestures that represent abstract concepts as physical objects or spaces (e.g., moving hands forward to represent the ``future'', or moving hands apart to show ``difference'').
    \item \textbf{Deictic:} Gestures that point to spatial locations, objects, or people (e.g., pointing down for ``here'', or pointing outward for ``them'').
    \item \textbf{Beat / Non-Semantic:} Rhythmic hand motions that align with speech prosody but carry no visual meaning. Words that primarily elicit beat gestures should be heavily penalized in your scoring.
\end{itemize}

\textbf{Instructions for each word:}
\begin{enumerate}
    \item \textbf{Score (1-5):} Assign a Gesturability Score from 1 to 5.
    \begin{itemize}
        \item 1 = Almost never accompanied by a semantic gesture (mostly beat or no gesture).
        \item 5 = Highly likely to consistently elicit a distinct, semantic gesture.
    \end{itemize}
    \item \textbf{Categorize:} Identify the primary gesture type it induces (Iconic, Metaphoric, or Deictic). If the score is 1 or 2, output ``Beat/None''.
    \item \textbf{Pattern:} Provide a short, 2-to-5 word phrase describing the core kinematic motion. (Write ``N/A'' if Beat/None).
    \item \textbf{Description:} Provide a detailed 1-to-2 sentence explanation of how the hands/body physically move to convey the word's semantics. (Write ``N/A'' if Beat/None).
\end{enumerate}

\textbf{Output Format:} Provide the results strictly in comma-separated values (CSV) format. Do not include any introductory or concluding text outside of the CSV. You must enclose the Pattern and Description fields in double quotes to prevent internal commas from breaking the CSV structure.

\textbf{Use the following header row:}\\
\texttt{Word,Gesturability Score,Gesture Category,Kinematic Pattern,Detailed Description}

\textbf{Examples of expected output:}\\
\texttt{"big",5,"Iconic","Hands move outward","The speaker typically starts with hands held neutrally and pushes the palms outward laterally to physically demonstrate a large spatial volume."}\\
\texttt{"however",1,"Beat/None","N/A","N/A"}
\end{tcolorbox}

\subsection{Pre-processing for stable gesture crops}
MultiVSR~\cite{multivsr} metadata provides the temporal boundaries of relevant speech segments. Unlike rigid face tracking, upper-body bounding boxes vary drastically with gesture articulation (e.g., wide arm movements). Consequently, standard frame-by-frame cropping and resizing introduces severe aspect-ratio fluctuations and motion jitter. To address this, we design a stabilization pipeline to preserve true gesture dynamics. For each segment, we extract temporally consistent person tracks using YOLOv9~\cite{wang2024yolov9} and an IoU-based greedy tracker. To prevent jitter, rather than cropping frames independently, we compute a single global crop per track from the aggregated bounding boxes and apply it uniformly. Since detections may include full bodies, we apply a leg-visibility heuristic to remove lower-body regions, yielding temporally stable, upper-body crops.

\subsection{Word recognition annotation}
The goal of this annotation is to obtain semantic vs.\ non-semantic labels for word clips. The input to the annotators is a set of 161,828 candidate clips, where a clip is a 4-second video containing the target word. We recruit annotators and train them using curated examples and detailed written guidelines. A central instruction is to adopt a ``conservative positive labeling'' strategy:
\begin{itemize}
    \item Mark a sample as \textit{Good} only if there is a \textbf{clearly identifiable gesture that depicts the target word} in a way that would be recognizable even without audio.
    \item Mark beat and random gestures without depictive content as \textit{Bad}.
    \item Mark gestures that are ambiguous, off-screen, or unrelated to the target word as \textit{Bad}.
\end{itemize}

\begin{figure}
    \centering
    \includegraphics[width=\linewidth]{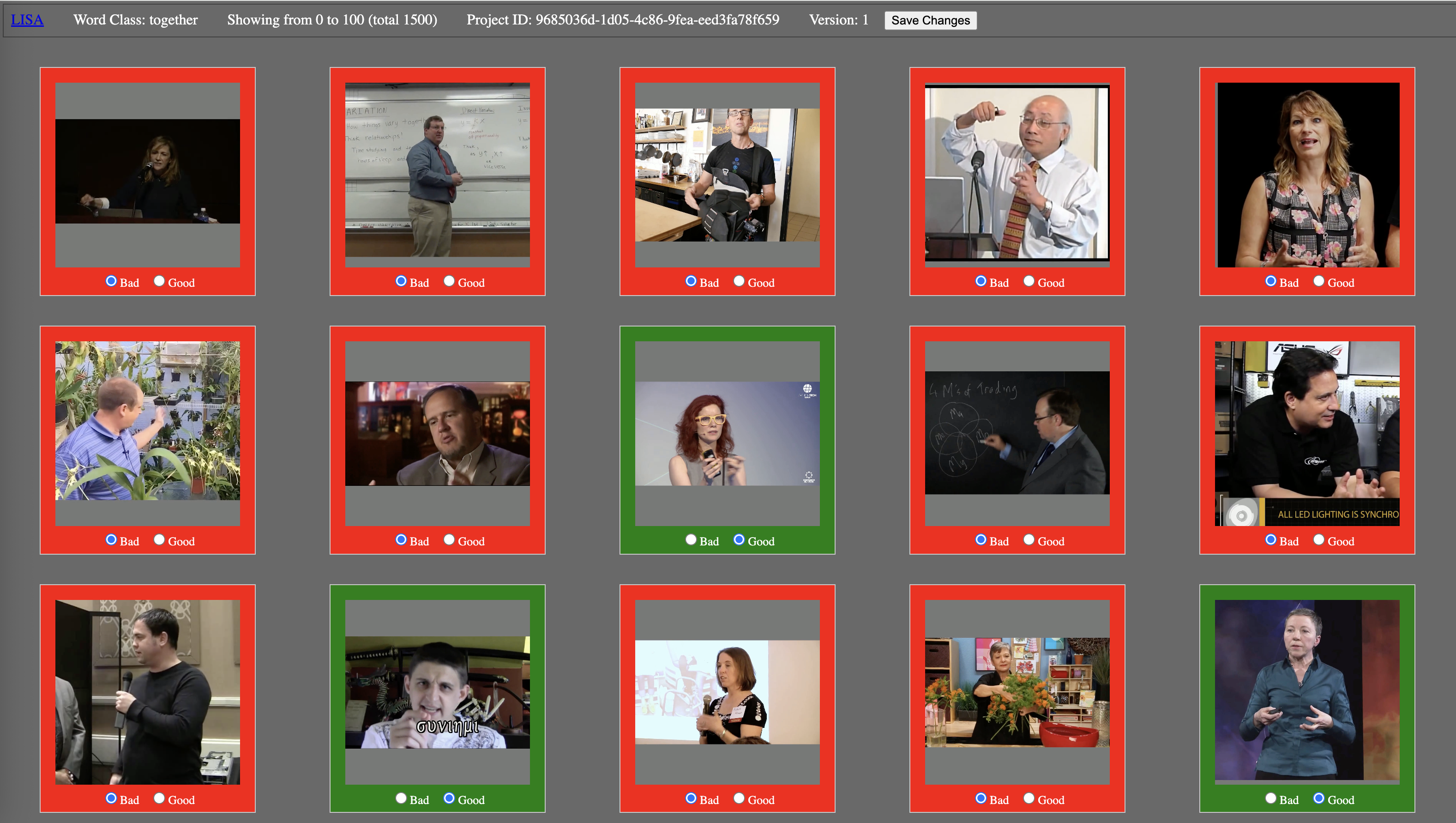}
    \caption{\textbf{Stage-1: Semantic Gesture Annotation Interface.} We developed a custom web-based tool built upon the VIA framework to efficiently collect binary labels. For a given target word (e.g., `together', shown at the top left), annotators are presented with a grid of candidate video clips. Annotators can play the audio and video for each clip, subsequently selecting ``Good'' if a clear, visually depictive semantic gesture is present, or ``Bad'' if the motion is non-semantic or absent. Hovering over each video also plays the audio as well, further assisting the annotator to judge and assign the correct label. After this stage, we obtain the set of semantic clips, and also the word label for each of the semantic clips. 
    }
    \label{fig:annotation_recog}
\end{figure}

We develop a custom in-house annotation interface built on VIA~\cite{dutta2019vgg}. Fig~\ref{fig:annotation_recog} shows a sample page from the word recognition annotation tool. Each page contains up to 100 samples arranged in a grid format, and there may be multiple pages for each word depending on the total number of instances. Each clip loops continuously, and audio playback is triggered on hover to facilitate inspection of the spoken context. By default, all clips are initialized as \textit{Bad} to minimize errors in the semantic annotations. Annotators assign a binary label indicating whether the target word is accompanied by a semantically meaningful gesture in the clip.

\subsection{Word localization annotation}
The input to this stage consists of the set of clips labeled as semantic gestures during the preceding recognition annotation. Specifically, we obtain 19,364 clips marked as \textit{Good}, which are known to contain semantic gestures and are passed to the localization stage. The annotators' task is to identify the temporal boundaries of the gesture by marking its start and end times. This task is more challenging than recognition, as it requires careful inspection of the entire clip, often through repeated viewing, to precisely determine the gesture boundaries corresponding to the target word. The instructions given to the annotators are as follows:
\begin{itemize}
    \item The start time should capture the moment just before the gesture begins (i.e., when the speaker leaves the rest position) or transitions from a previous gesture.
    \item The end time should correspond to the point at which the gesture is fully completed, either when the speaker returns to rest or transitions into another gesture.
\end{itemize}

\begin{figure}
    \centering
    \includegraphics[width=\linewidth]{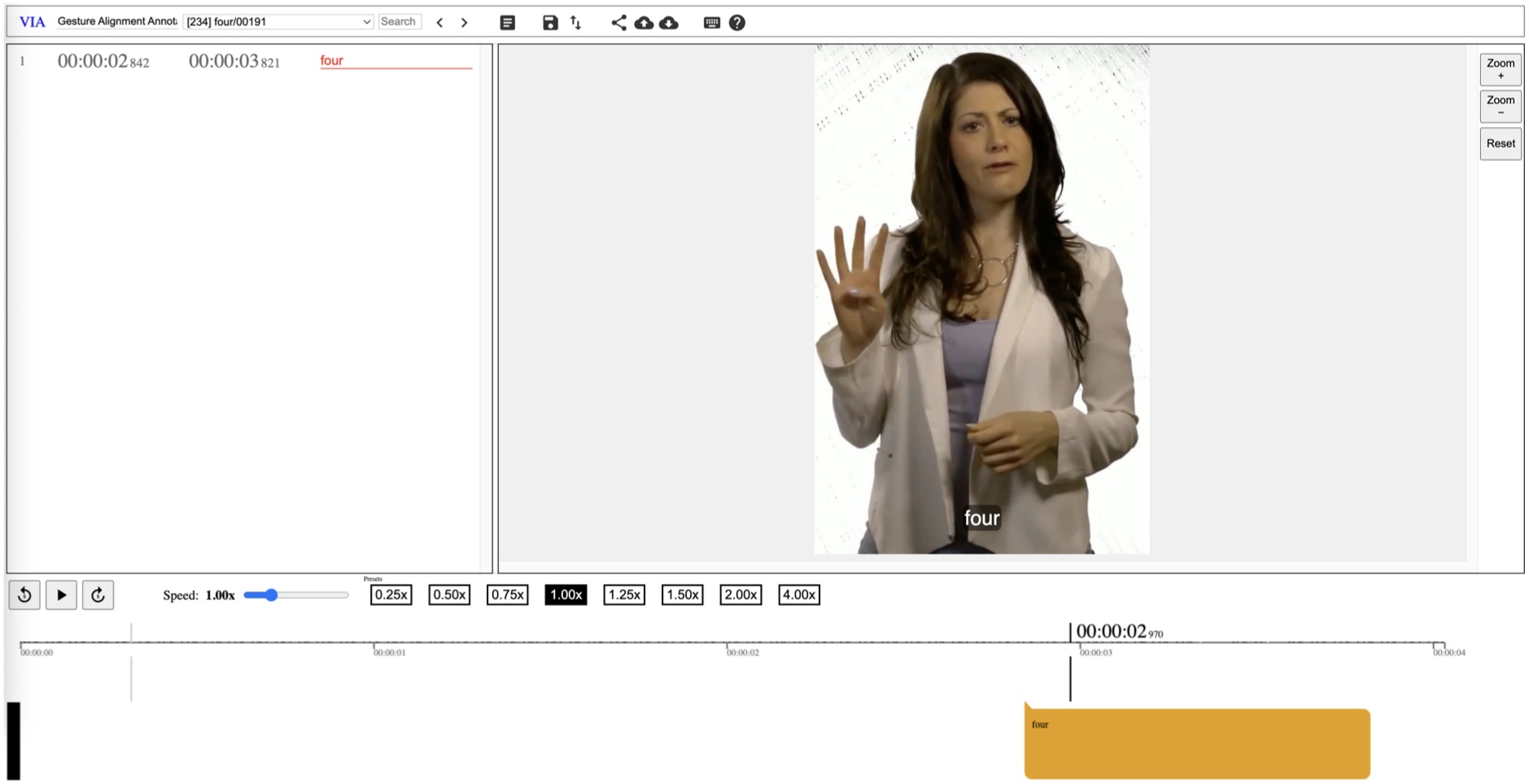}
    \caption{\textbf{Stage-2: Temporal Boundary Annotation Interface.} Using an adapted VIA interface, annotators precisely localize the start and end frames of a semantic gesture. For clips verified as positive in Stage-1, annotators utilize fine-grained playback speed controls and a draggable timeline to define the exact temporal envelope of the physical movement (e.g., localizing the gesture for the word `four' as shown here by the orange block). The interface pre-populates the temporal boundaries based on speech-timestamps for the target word, allowing annotators to subsequently adjust them to match the observed gesture after reviewing the video.
    }
    \label{fig:annotation_loc}
\end{figure}

We customize the annotation interface based on VIA~\cite{dutta2019vgg}. A sample annotation page is shown in Fig~\ref{fig:annotation_loc}. Each page displays a single video clip, allowing annotators to replay it multiple times, zoom in for clearer visual inspection, and adjust the playback speed as needed. 

We obtain 17,340 video clips as the output of the localization annotation, which constitute the semantic subset of the GRW dataset. To ensure data quality, all clips in this set are independently reviewed and validated by one of the authors.

\subsection{Annotator details}
We hired a total of 40 annotators through a professional annotation company. Prior to beginning the task, all annotators completed mandatory training sessions designed to familiarize them with the annotation guidelines, task objectives, and quality expectations. To ensure a consistent understanding of the criteria, annotators were required to complete a set of qualification test cases, which were reviewed before granting access to the full dataset. Throughout the annotation process, quality was monitored through periodic checks and feedback. All annotators were compensated at rates above the industry average to ensure fair pay and to promote careful, high-quality work.

\section{Model Details}

We provide more details on the proposed models and the training configuration. 

\subsection{Details on SHuBERT}

As discussed in Section~6.1 of the main paper, SHuBERT~\cite{gueuwou2025shubert} serves as the gesture backbone. Given an input video $V \in \mathbb{R}^{T \times H \times W \times 3}$, the left and right hands are cropped and processed using a DINO-v2 encoder fine-tuned on hand data and subsequently frozen, producing per-frame features of size $T \times 384$ for each hand. In parallel, body keypoints are extracted to obtain a $T \times 14$ representation. SHuBERT can additionally incorporate facial features when available; however, in our setup, facial cues are omitted to focus exclusively on hand-driven information. The modality-specific features are independently projected to a shared space, concatenated, and further projected to $T \times 768$ before being fed into a $12$-layer Transformer encoder, whose contextualized outputs constitute our final gesture backbone features, $f_{\text{multi}} \in \mathbb{R}^{L \times T \times d_{in}}$, where $L=12, d_{in}=768$. 

The next step in our setup is to aggregate the $12$ Transformer layers from $f_{multi}$ to obtain a unified representation $f \in \mathbb{R}^{T \times d_{in}}$. Several strategies can be considered for this aggregation: (i) using only the final layer's output, (ii) averaging representations across all layers, or (iii) learning a weighted combination of the layer-wise features. Empirically, we found the third approach to yield the best performance. Specifically, we have a fully-connected layer that inputs $d_{in}$-dimensional feature vectors of $f_{multi} \in \mathbb{R}^{L \times T \times d_{in}}$ and outputs a scalar for each of the feature vector: $\mathbb{R}^{L \times T \times 1}$. We get per-layer scores after averaging the feature-level scalars across the temporal dimension $T$: $\mathbb{R}^{L \times 1}$. These layer-wise scores are softmax-normalized over $L$, and a weighted sum across the layers to reduce $f_{\text{multi}} \in \mathbb{R}^{L \times T \times d_{in}}$ to $f \in \mathbb{R}^{T \times d_{in}}$.

\subsection{Architecture}
In Table~\ref{tab:arch}, we provide detailed architecture description for our two models: (i) semantic classification and (ii) word recognition and localization. The code and trained models have been released to support future research.

\begin{table*}
\centering
\caption{\textbf{Detailed Architectural Specifications.} Layer-by-layer breakdown of the input and output tensor shapes for our two proposed task-specific models. Notation: $T$ represents the number of frames in the extended temporal context, $Q$ denotes the number of frames in the shorter query interval (target clip), $L$ is the number of intermediate hidden layers extracted from the pre-trained gesture backbone, and $N$ is the total number of semantic word classes for the recognition task. Both models process $224 \times 224$ spatial RGB crops.}
\label{tab:arch}
\setlength{\tabcolsep}{4pt}
\begin{tabular}{|l|l|l|l|}
\hline
\textbf{Model} & \textbf{Layer/Module} & \textbf{Input Shape} & \textbf{Output Shape} \\
\hline

\rowcolor{Gray}
\multicolumn{4}{|l|}{\textbf{Semantic Classification}} \\
\hline
& Gesture backbone & 3 × T × 224 x 224 & L × T × 768 \\
& Layerwise feature aggregation & L × T × 768 & T × 768\\
\cline{2-4}
& Projection MLP & & \\
& - Linear & T × 768 & T × 512 \\
& - LayerNorm & T × 512 & T × 512 \\
& - ReLU & T × 512 & T × 512 \\
& - Linear & T × 512 & T × 512 \\
\cline{2-4}
& Transformer Encoder & & \\
& - Self-Attention & T × 512 & T × 512 \\
& - Feed Forward & T × 512 & T × 512 \\
\cline{2-4}
& Transformer Decoder & & \\
& - Self-Attention & \textit{Q} × 512 & \textit{Q} × 512 \\
& - Cross-Attention & \textit{Q} × 512 | T × 512 & \textit{Q} × 512 \\
& - Feed Forward & \textit{Q} × 512 & \textit{Q} × 512 \\
\cline{2-4}
& Binary Classification & & \\
& - Mean Pooling & \textit{Q} × 512 & 512 \\
& - Output Projection & 512 & 512 \\
& - Classification & 512 & 1 \\
\hline

\rowcolor{Gray}
\multicolumn{4}{|l|}{\textbf{Word Recognition and Localization}} \\
\hline
& Gesture backbone & 3 × \textit{Q} × 224 x 224 & L × \textit{Q} × 768 \\
& Layerwise feature aggregation & L × \textit{Q} × 768 & \textit{Q} × 768\\
\cline{2-4}
& Projection MLP & & \\
& - Linear & \textit{Q} × 768 & \textit{Q} × 768 \\
& - LayerNorm & \textit{Q} × 768 & \textit{Q} × 768 \\
& - ReLU & \textit{Q} × 768 & \textit{Q} × 768 \\
& - Linear & \textit{Q} × 512 & \textit{Q} × 768 \\
\cline{2-4}
& Transformer Encoder & & \\
& - Self-Attention & \textit{Q} × 768 & \textit{Q} × 768 \\
& - Feed Forward & \textit{Q} × 768 & \textit{Q} × 768 \\
\cline{2-4}
& Localization Head & & \\
& - MLP & \textit{Q} × 768 & \textit{Q} × 768 \\
& - Localization & \textit{Q} × 768 & \textit{Q}\\
\cline{2-4}
& Classification Head & & \\
& - Weighted Average & \textit{Q} × 768 & 768 \\
& - Output Projection & 768 & 768 \\
& - Classification & 768 & \textit{N} \\  
\hline
\end{tabular}
\end{table*} 

\subsection{Training hyper-parameters}
All our models are implemented in PyTorch. We train both models until validation loss does not improve for 5 epochs. The hyper-parameters for each of the individual models are stated below. 

\subsubsection{Semantic gesture classification.} We use 3 transformer encoder layers and 3 transformer decoder layers. We use an initial learning rate of $1e^{-4}$ and decrease it by a factor of 5 every time the validation loss plateaus for 2 epochs. Since the data is heavily class imbalanced (15,340 positive semantic gesture samples and 120,163 negative samples), we use class-balanced sampling to upsample positive class data in each batch, and additionally apply a class-weighting to the loss (assigns higher weight to positive samples). We also apply the following feature-level temporal augmentation during training: (i) dropping frames, (ii) shifting frames, and (iii) changing the speed of video by interpolating frames. We use a batch size of 128 and train the model with AdamW optimizer on a single GPU. 

\subsubsection{Word recognition and localization.} We use 6 transformer encoder layers. We use an initial learning rate of $1e^{-4}$ and decrease it by a factor of 5 every time the validation loss plateaus for 2 epochs. We use a batch size of 128 and train the model with AdamW optimizer on a single GPU. 

\clearpage

\section{Gemini Prompts for Evalutation}
We provide detailed prompts that are used to evaluate Gemini-3.5 on the gesture tasks. We run the following prompts for all the samples in our test set for the both the tasks.

\begin{tcolorbox}[fontupper=\scriptsize, title=Prompt for Semantic Gesture Classification, colback=gray!5!white, colframe=gray!75!black]
\textbf{Role:} You are an expert in gesture recognition.

\textbf{Task:} Analyze the provided short video clip and determine whether the person makes a clear iconic / deictic / metaphoric hand gesture.\\

\textbf{Guidelines:}
\begin{itemize}
    \item Focus only on hand movements, shapes, and orientation changes. Ignore body posture, facial expressions, or background.
    \item A gesture may occur anytime in the duration of the input video.
    \item Random continuous hand movements or beat gestures cannot be considered as ``Good''.
    \item If there is no clear hand gesture present in the full video, output ``No gesture found''.
    \item Be precise and confident only if the gesture is clear and is depictive of a word. Example: arms widening to indicate ``massive'', hands moving in a circular manner to indicate ``round''. 
\end{itemize}

\textbf{Steps to Follow:}
\begin{itemize}
    \item Carefully examine the hand movements throughout the video.
    \item Check if any movement could potentially strongly correspond to semantic gestures, i.e., iconic / deictic / metaphoric gestures.
    \item The following are different possibilities:
    \begin{enumerate}
        \item No semantic gestures present in the video.
        \item Semantic gesture may be present, but is not clear and/or is not fully visible.
        \item Semantic gesture present, and is clearly visible.
    \end{enumerate}
    \item Make a decision: ``Yes'' (gesture found) or ``No'' (gesture not found). The decision will be ``Yes'' only if category 3 occurs as explained above.
    \item Give the confidence score between 0 and 1, 0 being the least confident and 1 being the most confident.
    \item Provide a short reasoning.
\end{itemize}

\textbf{Output Format:}\\
\texttt{Gesture present: <Yes or No>}\\
\texttt{Confidence: <0.0 to 1.0>}\\
\texttt{Reasoning: <Brief explanation of the decision>}
\end{tcolorbox}

\begin{tcolorbox}[fontupper=\scriptsize, title=Prompt for Word Recognition and Localization, colback=gray!5!white, colframe=gray!75!black]
\textbf{Role:} You are an expert in gesture recognition and localization.

\textbf{Task:} You will be given a short video clip with a person speaking. Your job is to:
\begin{enumerate}
    \item Analyze all the frames of the video together and produce one final gesture description that best represents what the hands are gesturing throughout these frames.
    \item Based on the description, rank a given set of 100 possible word classes with confidence scores (between 0 and 1) such that they sum to approximately 1.0.
    \item Predict the start and end time of the gesture in seconds (gesture boundary) for the recognized target word.
\end{enumerate}

\textbf{Classes:} bye, below, push, move, direction, specific, together, whole, hello, straight, above, switch, five, turn, expand, throw, open, huge, raise, large, tiny, long, mix, circle, no, two, three, entire, top, four, lower, full, press, small, grab, grow, separate, wrap, wide, roll, block, combine, high, big, round, increase, flip, force, close, develop, lift, narrow, hold, broad, connect, begin, global, build, reduce, process, spin, front, focus, cross, layer, count, wave, curve, short, shake, loop, height, strong, stop, us, run, little, stretch, track, zoom, condense, evolve, around, interaction, twist, transition, back, less, walk, absorb, quick, bottom, her, descend, peak, join, perfect, transform, decrease, stack.\\

\textbf{Step 1: Gesture Description Guidelines}
\begin{itemize}
    \item Look at the hand shapes, positions, and movement across the frames.
    \item Focus only on hand gestures, ignore background or person details (e.g., clothing, gender, facial expressions).
    \item Summarize the gesture as a single concise description, not frame-by-frame.
    \item If the gesture evolves across frames, capture that motion pattern in the final description (e.g., ``raising hand then pointing forward'').
    \item If uncertain, provide the most plausible interpretation based on the sequence.
    \item Use short, clear, natural language (e.g., ``hands forming a circle'', ``index finger pointing to the right'').
\end{itemize}

\textbf{Step 2: Gesture Recognition Guidelines}
\begin{itemize}
    \item Carefully analyze the gesture description and reason about its likely meaning.
    \item Assign a confidence score between 0 and 1 for each of the 100 classes based on how likely it matches the gesture.
    \item The most likely class gets the highest confidence, followed by others in descending order.
    \item If none of the classes match perfectly, rank and distribute class-wise probabilities regardless.
    \item Ensure the sum of all confidence scores approximately equals 1.0.
\end{itemize}

\textbf{Step 3: Gesture Localization Guidelines}
\begin{itemize}
    \item Carefully analyze the gesture description and the output top recognized word (top-confident word above).
    \item Focus only on this target word, this is the word to be localized.
    \item Determine the start time when the gesture for this particular target word begins (e.g., hand moves or shapes start forming) or transitions from a previous gesture. The start time should capture the moment just before the gesture begins (i.e., when the speaker leaves the rest position).
    \item Determine the end time when the gesture for this particular target word completes (e.g., hand returns to neutral or stops moving). The end time can either be when the speaker returns to rest position or transitions into another gesture.
\end{itemize}

\textbf{Output Format:}\\
\texttt{Gesture description: <final description>}\\
\texttt{Ranked classes with confidence:}\\
\texttt{\{}\\
\texttt{\ \ \ \ <class-1> - <confidence>}\\
\texttt{\ \ \ \ <class-2> - <confidence>}\\
\texttt{\ \ \ \ \ \ \ \ \ \ \ \ ...}\\
\texttt{\ \ \ \ <class-100> - <confidence>}\\
\texttt{\}}\\
\texttt{Gesture start: <start time in seconds>}\\
\texttt{Gesture end: <end time in seconds>}\\
\texttt{Localization confidence: <0.0 to 1.0>}
\end{tcolorbox}

\section{Failure cases}

\begin{figure}
    \centering
    \includegraphics[width=\linewidth]{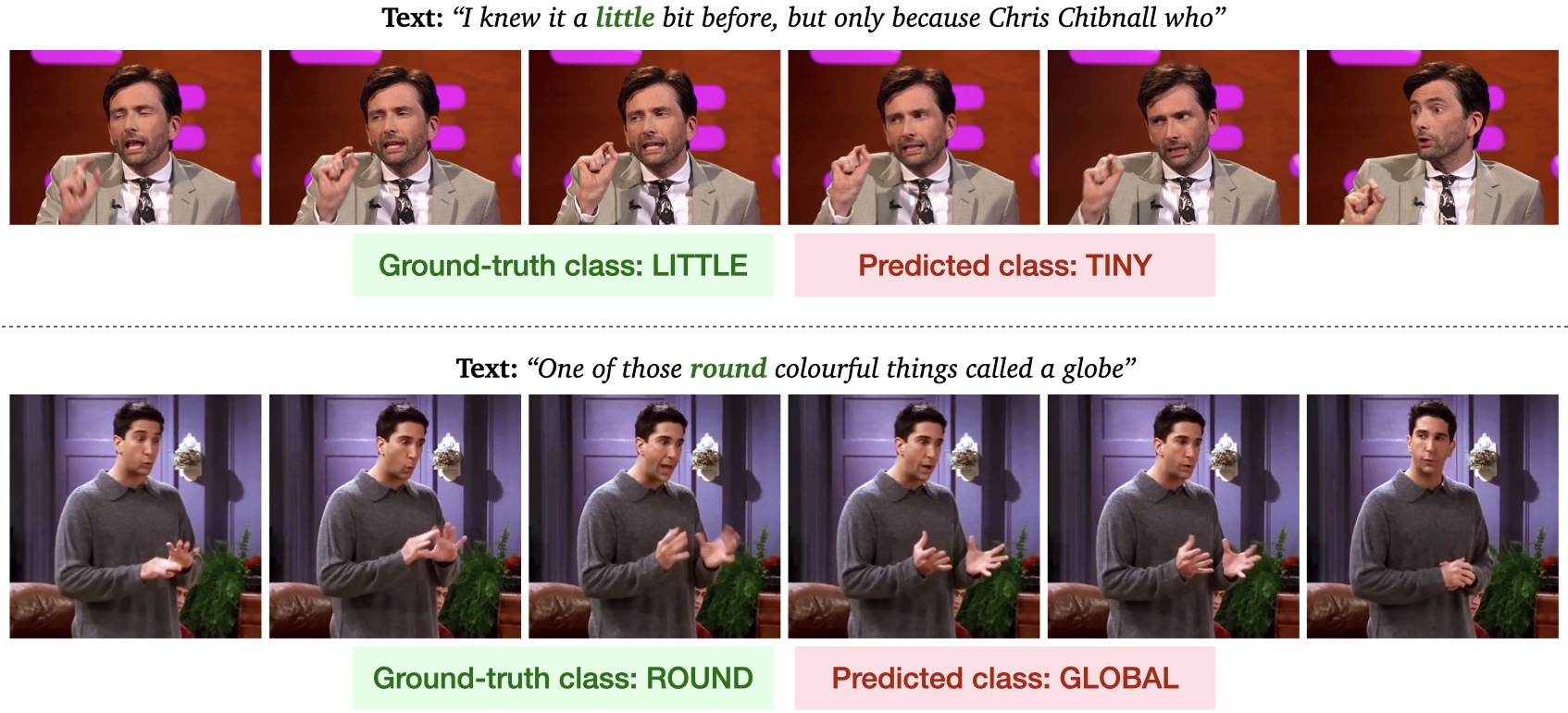}
    \caption{\textbf{Analysis of typical failure cases.} The word recognition model occasionally misclassifies gestures into semantically and visually synonymous categories. \textbf{(Top)} A classic finger-pinching motion results in a prediction of ``tiny'' rather than the ground-truth ``little''. \textbf{(Bottom)} The model predicts ``global'' instead of the ground-truth ``round'' as the speaker gestures a circular hand motion. These examples highlight the inherent ambiguity in word-level gesture recognition, where different lexical items share nearly identical physical expressions.}
    \label{fig:failure}
\end{figure}

An analysis of the word recognition model's failure modes highlights the natural ambiguity of human co-speech gesticulation. As illustrated in Fig.~\ref{fig:failure}, misclassifications typically occur between words that share nearly identical physical expressions. Rather than producing random predictions, the model often selects valid synonyms, such as predicting ``tiny'' instead of ``little'', or ``global'' instead of ``round''. These semantically related predictions demonstrate that the model successfully captures the underlying visual meaning, even if it misses the exact lexical ground truth. This also underscores why top-$k$ metrics, where $k=5, 10$ are essential for evaluating unconstrained gesture recognition.

\end{document}